\documentclass[acmlarge]{acmart}
\usepackage{graphicx} %Loading the package
\graphicspath{{figures/}} %Setting the graphicspath
\usepackage{tabularx}

\usepackage{tikz}
\usepackage{comment}
\usepackage{amsmath} % define this before the line numbering.
\usepackage{color}
\usepackage{hyperref}

\usepackage{balance}
\usepackage{url}
\usepackage{multirow}

\newcommand{\quoted}[1]{``#1''}
\newcommand{\mycomment}[1]{}

\usepackage{array}
\usepackage[normalem]{ulem}

\AtBeginDocument{%
  \providecommand\BibTeX{{%
    \normalfont B\kern-0.5em{\scshape i\kern-0.25em b}\kern-0.8em\TeX}}}

\setcopyright{acmcopyright}
\copyrightyear{2020}
\acmYear{2020}
\acmDOI{10.1145/1122445.1122456}

\acmJournal{JOCCH}
\acmVolume{37}
\acmNumber{4}
\acmArticle{111}
\acmMonth{8}

\begin{document}

\title{A Data Set and a Convolutional Model for Iconography Classification in Paintings}

\author{Federico Milani}
\email{federico.milani@polimi.it}
\orcid{0000-0003-2700-2228}
\author{Piero Fraternali}
\email{piero.fraternali@polimi.it}
\orcid{0000-0002-6945-2625}
\affiliation{%
  \institution{Politecnico di Milano}
  \streetaddress{Piazza Leonardo da Vinci, 32}
  \city{Milan}
  \state{Italy}
  \postcode{22100}
}

\renewcommand{\shortauthors}{Milani, et al.}
\renewcommand\tabularxcolumn[1]{m{#1}}

\begin{abstract}
Iconography in art is the discipline that studies the visual content of artworks to determine their motifs and themes and to characterize the way these are represented. It is a subject of active research  for a variety of purposes, including the interpretation of meaning,  the investigation of the origin and diffusion in time and space of representations, and the study of influences across artists and art works. With the proliferation of digital archives of art images, the possibility arises of applying Computer Vision techniques to the  analysis of art images at an unprecedented scale, which may support iconography research  and education. In this paper we introduce a novel paintings data set for iconography classification and  present the   quantitative and qualitative results of applying a Convolutional Neural Network (CNN) classifier to the   recognition of the iconography of artworks. The proposed classifier achieves good performances (71.17\% Precision, 70.89\% Recall, 70.25\% F1-Score and 72.73\% Average Precision) in the task of identifying saints in Christian religious paintings, a task made difficult by the presence of classes with very similar visual features. Qualitative analysis of the results shows that the CNN focuses on the traditional iconic motifs that characterize the representation of each saint and exploits such hints to attain correct identification. The ultimate goal of our work is to enable the automatic extraction, decomposition,  and comparison of iconography elements to support iconographic studies and automatic art work annotation.
\end{abstract}




\maketitle

\section{Introduction}\label{sec:intro}

\textit{Iconography is that branch of the history of art which concerns itself with the subject matter or meaning of works of art, as opposed to their form} \cite{panofsky39}. Iconography focuses  on the subject  of an artwork and on the way it is represented. The iconographic analysis of an artwork determines the subject portrayed in the image (e.g., the crucifixion or the visitation of the magi) and characterizes the way in which such a theme is represented, e.g, the number of subjects or objects that constitute the representation, their disposition and mutual relations, and ultimately their possible symbolic meaning. 
Iconographic studies have a fundamental role in art history, because iconography is a cultural marker, which can be used to support the identification of  the production period and to delimit the region of provenance of an artwork, and to understand the intention of the commissioner or of the author. 
The popularization of digital art images, which are now available in massive numbers in open data sets \cite{mensink2014rijksmuseum,stefanini2019artpedia}, offers new possibilities to art history and iconography studies. Researchers have now at their disposition a wealth of images, usable to investigate the evolution of iconography motifs  and themes in time and space, to assess the influences across  artworks and authors, and even to conduct cross-media studies, which e.g., may reconnect a specific iconography to its literary source.  A revealing signal of the increasing importance of IT-supported iconography studies is the effort of standardizing the identification of the content of images by means of a single agreed-upon iconography taxonomy. The most prominent effort in this direction is the \textit{Iconclass} system \cite{couprie1983iconclass}, which provides over 28k classification types for ten top-level categories of images and is increasingly used by art historians and collection curators to label art images in a way that facilitates search and comparison.
However, both the manual creation and the analysis of  labelled art images data sets is  a non-trivial and time consuming task. Iconographic classes are numerous,  may denote complex content made of multiple interrelated motifs, can present themselves in multiple variants, and can be sub-structured into elements whose identification and characterization may be important too.

Computer Vision (CV) methods may be of great help in supporting the production and the exploitation  of iconography-oriented art images data sets: 1) given an image, the set of iconography classes potentially denoting its content could be retrieved and ranked by likelihood, to support the semi-automatic labeling of the data set; 2) the regions of the image where the core elements of the iconography class are present could be detected, supporting the morphological analysis of the iconography and the detection of specific variants    or of important sub-elements; 3) a data set of art images could be clustered based on the iconography class of artworks, the specific variant they express, and the presence or position of specific sub-elements. 
Machine Learning (ML) and CV techniques are the natural candidates to support the iconography analysis of art images. 
In its \quoted{simplest} form, this task can be formulated as a classification problem. Classification of natural images by their content is now considered a solved CV task. Conversely, classifying artwork iconography is much less investigated and still presents non trivial challenges: iconographic classes are numerous and structurally complex, high-quality annotated data sets are not commonly available, and the art images themselves are visually more homogeneous than natural images which makes their discrimination difficult. 
Training a CV component for the classification of art iconography requires creating a data set annotated at the image level with the iconography class(es) portrayed in the artwork.
Such labelling could be semi-automated, e.g., by exploiting keywords and metadata associated with each image, but  expert judgement is still fundamental to provide missing labels  or correct wrong annotations. 

The more interesting problem of characterizing the image regions where the constituents of an iconography class are found requires addressing the much harder problem of multi-instance object detection \cite{lin2014microsoft}.
The definition of a high quality  data set, supporting the training of a component for the detection of the constituents of an iconography class is far from trivial, because it would require an expert to  manually annotate at the pixel level  a sufficient number of examples for a very high number of classes (28k); if the analysis of specific objects or parts of an iconography (e.g., the number of nails in a crucifixion scene\footnote{Three nails crucifixions started replacing four nails ones in the middle of the XIII century; the number of nails helps researchers  delimit the production period of an artwork \cite{heinz71}}, the position and number of angels in a annunciation scene, etc.)  should be supported, the effort would be even greater. 
%These difficulties make the automatic characterization of art image iconography a perfect case for weakly supervised methods, which alleviate the burden of fine-grain data set labeling by exploiting the inner layer of a Deep Learning (DL) classifier trained on whole images \cite{zhou2016learning}.

In this paper, we address the problem of iconography analysis  by tackling the initial task of iconography class identification in images of paintings. As a case study, we consider iconography classes associated with saints in christian religious art. 
The contributions of the paper can be summarized as follows:

\begin{itemize}
    \item We introduce a novel data set, called \textit{ArtDL},  for iconography classification, which consists of 42k  annotated images pertaining to 19 classes (10 of which are \quoted{long tail} having less than 1000 annotated samples). Each class denotes the presence of a specific character (e.g., Virgin Mary, Saint Sebastian, Saint Anthony of Padua, Saint Mary Magdalene, Saint Francis of Assisi, Saint Jerome, Saint Peter, Saint Paul, Saint Dominic, Saint John the Baptist).
    \item We develop a CNN classifier for the task of iconography class recognition. We address the problem of the limited amount of labeled data in the art iconography domain by applying transfer learning \cite{pan2009survey}, to re-use common knowledge extracted from a model pre-trained on ImageNet \cite{deng2009imagenet}. In the best experiment, with the 10 most discriminative classes, the classifier achieves 72\% average precision and 70\% F1 score, with 71\% precision at 70\% recall.
    \item   We analyze the output of the classifiers qualitatively   by exploiting visual understanding and interpretability techniques (specifically Class Attention Maps -- CAMs \cite{zhou2016learning})  to identify the representative image regions where the classifier  focuses its attention. We also illustrate examples of confusion among classes, mostly in cases of strong visual similarity, to  better highlight the hardness of the task.
    \item We publish both the data set and the trained CNN model at the address: \texttt{http://www.artdl.org}
\end{itemize}

The rest of the paper is organized as follows:
Section \ref{sec:relwork} overviews the related work on the IT-supported tasks and methods for artwork image analysis and on the main data sets that enable research in such a  field; it also briefly reviews the essential results about model interpretability, a key technique to support the qualitative analysis of our results.  
Section \ref{sec:method} introduces the ArtDL data set for saint iconography in artworks and the proposed architecture for addressing the saint iconography  classification task;
Section \ref{sec:results} presents the results of applying the defined architecture to the ArtDL data set from both a quantitative and a qualitative point of view;
Section \ref{sec:concl} concludes and provides an outlook on the future work.

\section{Related Work}\label{sec:relwork}

\subsection{Computer Vision and Deep Learning for Artwork Image Analysis}
% folder to papers https://drive.google.com/open?id=1xNATZ285q7x-xe7CPwzt--DkyfjKinSC

The increased availability of visual art content in digital  format has spawned the research interest in methods  for supporting digital humanities studies and  cultural heritage asset management.   In this section we report the most relevant research results addressing the different tasks of artwork image analysis and describe the main data sets created to support such efforts. 

%The difficulty of automating the analysis of  artworks, with respect to photographs, is that their  style and interpretation are influenced by cultural factors, such as the author's personality and background,  the production period,  and the region of provenance. An overview of techniques and applications can be found in \cite{bartolini2003applications,brachmann2017computational}.% including not only paintings, but sculptures, 
%to read, i think it is also about style: \cite{florea2016pandora}
%CV and more recently DL techniques have been successfully applied  to the analysis of images and videos for a variety of tasks such as classification, object detection, and instance segmentation in multiple domains \cite{guo2016deep}.

\subsubsection{Artwork analysis tasks}

The \textit{artwork classification by style} is one of the first areas of study. The early work \cite{zujovic2009classifying} explored the automatic categorization of paintings by  genre (Impressionism,  Cubism, Impressionism, Pop Art and Realism)
using a data set of $\sim$350  images from different sources (Google, Artlex, CARLI Digital Collections). The proposed method extracted multiple image features and   used them to train different classifiers: Naive Bayes, K-Nearest-Neighbours,   Support Vector Machines, Decision Trees and Artificial Neural Networks. The results demonstrated the feasibility of the task, even with such a small data set. The work \cite{gao2015adaptive} proposed a method for style recognition based on Adaptive Sparse Analysis that exploits Discrete Cosine Transform to learn a dictionary of distinctive painting characteristics. The experiments spanned  authentication, stylometry and classification tasks and showcased promising results.  More recent approaches adopt CNNs for  feature extraction, replacing the design of hand crafted features.  In  \cite{karayev2013recognizing} the authors compare linear classifiers based on hand-crafted features and  on features  obtained using  AlexNet \cite{krizhevsky2012imagenet} pre-trained with ImageNet \cite{deng2009imagenet}; the assessment exploits a data set of around 85,000 images labeled with 25 different art styles. The same approach is used in \cite{tan2016ceci}, where the features obtained by a variation of the AlexNet CNN  are used to train an SVM, showing that automatically extracted features can outperform classical image descriptors. Other examples of style image classification can be found in \cite{bar2014classification,lecoutre2017recognizing,florea2017domain, chen2017multi}.

Artists have their own imprint that makes experts recognize their work among others with the same style. The \textit{author identification task}  has been investigated in several studies   \cite{keren2002painter,mensink2014rijksmuseum}, with motivations that range from the automatic cataloguing of unlabeled works to the identification of forgery. The authors of \cite{shamir2012computer} proposed an automatic method to analyze paintings of several artists and schools and to group them by artistic movement. The generated phylogenies highlight similarities and influential links in agreement with art historians. Johnson et al. \cite{johnson2008image} proposed an approach to classify paintings by author, which relies among the other features on the segmentation of brushstrokes (using K-means clustering and edge extraction). A similar approach was employed in \cite{elgammal2018picasso}, where stroke segmentation was used for artist identification and forgery detection. Artists were asked to imitate different artworks (e.g., by Picasso)   to generate a test data set.  The work \cite{belhi2018towards} proposed a hierarchical multitask classification framework to automatically detect the characteristics  of an artwork and   produce   multiple metadata, such as author, year, genre and medium of paintings. 

The \textit{artwork content analysis task} focuses on automatically recognizing the  subject of artworks, by detecting the objects that appear in the image or by localizing their position  \cite{crowley2013gods,cai2015cross,gonthier2018weakly,shen2019discovering}. 
Applications include image retrieval and iconography labeling for asset management. In \cite{crowley2014state} the authors introduce the Paintings Dataset of 200,000+  British paintings annotated with ten classes (e.g., bird, boat, chair). The objective is to predict the object classes that appear in an input image.  The adopted approach consists of training a model on the PASCAL VOC \cite{everingham2015pascal} data set and then evaluating how well the trained model  performs when  moved from natural images to art paintings. This technique is  known as Transfer Learning (TL) \cite{pan2009survey} and is  frequently applied to art image analysis due to the absence of art-specific  models trained on large data sets. Several studies evaluate the transferability of previous knowledge to the art domain. The work of \cite{sabatelli2018deep} investigates the behavior of CNNs  pre-trained on a  different domain and fine-tuned with art images. The results show that fine-tuned models outperform models trained from scratch on art images. The same conclusions are obtained in \cite{cetinic2018fine} which studies transferability on different classification tasks: genre, artist, style recognition.
Another example of fine tuning is \cite{westlake2016detecting}, where the goal is to detect people in  artworks. The authors employ  the Fast R-CNN \cite{girshick2015fast} model pre-trained on ImageNet,  fine-tune it in the People-Art data set, and assess  performances on their own data set and on the Picasso data set,  previously used for detecting people in cubist artworks \cite{ginosar2014detecting}. 

More \textit{specific artwork analysis tasks}, such as sentiment detection or visual aesthetic analysis have also been investigated \cite{li2009aesthetic,amirshahi2014jenaesthetics,Alameda-Pineda_2016_CVPR,brachmann2017using, brachmann2017computational,kang2018method,cetinic2019deep}. For example,  \cite{cetinic2019deep} exploits CNNs to predict subjective aspects of human perception: aesthetic evaluation, evoked sentiment and memorability. The authors also analyze which features  of the images contribute the most to a given aspect and explore such findings in the context of art history.

Table \ref{tab:arttaskperf} surveys the principal results in the field of artwork image  analysis. 
The table lists   recent  works in order of appearance,  specifies the task they address, the reported performance results, and the data set(s) used in the evaluation. We can observe a high variance of results depending on the data set and on the task, e.g. artist identification has an accuracy between 30.2\% and 92.9\% while style classification has an accuracy that ranges between 39.1\% and 84.4\%.

\begin{table}[tp]
\centering
\small
\begin{tabular}{|p{3.2cm}|c|p{3.8cm}|p{3.9cm}|}
\hline
\textbf{Task} & \textbf{Year} & \textbf{Data set} & \textbf{Performance \%} \\
\hline
\multirow{19}{*}{Artist identification} & 2014 & Painting-91 & \textbf{Accuracy} 53.1 \cite{painting91} \\ \cline{2-4}
& 2015 & Painting-91 & \textbf{Accuracy} 56.4 \cite{peng2015cross} \\ \cline{2-4}
& 2016 & Painting-91 & \textbf{Accuracy} 64.5 \cite{anwer2016combining} \\ \cline{2-4}
& 2016 & Wikiart & \textbf{Accuracy} 76.1 \cite{tan2016ceci} \\ \cline{2-4}
& 2016 & Wikiart & \textbf{Accuracy} 63.1 \cite{saleh2015large} \\ \cline{2-4}
& 2016 & Painting-91 & \textbf{Accuracy} 63.2 \cite{chu2016deep}\\ \cline{2-4}
& 2016 & Painting-91 & \textbf{Accuracy} 59.0 \cite{puthenputhussery2016color} \\ \cline{2-4}
& 2016 & Painting-91 & \textbf{Accuracy} 65.8 \cite{puthenputhussery2016sparse} \\ \cline{2-4}
& 2016 & Painting-91 & \textbf{Accuracy} 57.3 \cite{peng2016toward} \\ \cline{2-4}
& 2016 & Painting-91 & \textbf{Accuracy} 45.0 \cite{banerji2016painting} \\ \cline{2-4}
& 2017  & Art500k & \textbf{Accuracy} 30.2 \cite{mao2017deepart} \\ \cline{2-4}
& 2017 & Painting-91 & \textbf{Accuracy} 78.5 \cite{bianco2017large} \\ \cline{2-4}
& 2017 & Wikiart & \textbf{Accuracy} 81.9 \cite{huang2017fine} \\ \cline{2-4}
& 2018 & Wikiart & \textbf{Accuracy} 81.9 \cite{cetinic2018fine} \\ \cline{2-4}
& 2018 & Wikiart & \textbf{Accuracy} 64.3 \cite{chu2018image} \\ \cline{2-4}
& 2018 & OmniArt & \textbf{Accuracy} 71.1 \cite{strezoski2018omniart} \\ \cline{2-4}
& 2018 & Rijkmuseum & \textbf{Accuracy} 92.9 \cite{sabatelli2018deep} \\ \cline{2-4}
& 2019 & Multitask Painting 100k & \textbf{Accuracy} 56.5 \cite{Bianco2019MultitaskPC} \\ \cline{2-4}
& 2020 & Wikiart & \textbf{Accuracy} 86.7 \cite{zhong2020fine} \\ \hline
\multirow{4}{*}{Content analysis}  & 2012 & PRINTART &  \textbf{F1-Score} 38.0 $\pm$ 3.0 \cite{carneiro2012artistic} \\ \cline{2-4} 
& 2016 & Paintings Dataset & \textbf{AP} 68.5 \cite{crowley2016art} \\ \cline{2-4}
& 2018 & Iconart & \textbf{mAP} 59.2 $\pm$ 1.2 \cite{gonthier2018weakly} \\ \cline{2-4}
& 2020 & Iconart & \textbf{mAP} 69.2 $\pm$ 0.3 \cite{gonthier2020multiple} \\ \hline
\multirow{24}{*}{Style classification} & 2013 & Flickr Style & \textbf{Mean AP} 36.8 \cite{karayev2013recognizing} \\ \cline{2-4}
& 2013 & AVA Style & \textbf{Mean AP}  58.1 \cite{karayev2013recognizing}  \\ \cline{2-4}
& 2013 & Wikiart & \textbf{Mean AP}  47.3 \cite{karayev2013recognizing}  \\ \cline{2-4}
& 2014 & Painting-91 & \textbf{Accuracy} 62.2 \cite{painting91} \\ \cline{2-4}
& 2015 & Painting-91 & \textbf{Accuracy} 69.2 \cite{peng2015cross} \\ \cline{2-4}
& 2016 & Painting-91 & \textbf{Accuracy} 78.4 \cite{anwer2016combining} \\ \cline{2-4}
& 2016 & Wikiart & \textbf{Accuracy} 54.5 \cite{tan2016ceci} \\ \cline{2-4}
& 2016 & Wikiart & \textbf{Accuracy} 46.0 \cite{saleh2015large} \\ \cline{2-4}
& 2016 & Painting-91 & \textbf{Accuracy} 73.6 \cite{chu2016deep} \\ \cline{2-4}
& 2016 & Wikiart & \textbf{Accuracy} 58.2 \cite{chu2016deep} \\ \cline{2-4}
& 2016 & Pandora & \textbf{Accuracy} 54.7 \cite{florea2016pandora} \\ \cline{2-4}
& 2016 & Painting-91 & \textbf{Accuracy} 67.4 \cite{puthenputhussery2016color} \\ \cline{2-4}
& 2016 & Painting-91 & \textbf{Accuracy} 73.2 \cite{puthenputhussery2016sparse} \\ \cline{2-4}
& 2016 & Painting-91 & \textbf{Accuracy} 70.1 \cite{peng2016toward} \\ \cline{2-4}
& 2016 & Painting-91 & \textbf{Accuracy} 64.5 \cite{banerji2016painting} \\ \cline{2-4}
& 2017 & Painting-91 & \textbf{Accuracy} 84.4 \cite{bianco2017large} \\ \cline{2-4}
& 2017 & Wikiart & \textbf{Accuracy} 50.1 \cite{huang2017fine} \\ \cline{2-4}
& 2017  & Art500k & \textbf{Accuracy} 39.1 \cite{mao2017deepart} \\ \cline{2-4}
& 2018 & Wikiart & \textbf{Accuracy} 78.3 \cite{chu2018image} \\ \cline{2-4}
& 2018 & Wikiart & \textbf{Accuracy} 56.4 \cite{cetinic2018fine} \\ \cline{2-4}
& 2018 & Painting-91 & \textbf{Accuracy} 67.3 \cite{falomir2018categorizing} \\ \cline{2-4}
& 2018 & OmniArt & \textbf{Accuracy} 51.2 \cite{strezoski2018omniart} \\ \cline{2-4}
& 2019 & Multitask Painting 100k& \textbf{Accuracy} 57.2 \cite{Bianco2019MultitaskPC} \\ \cline{2-4}
& 2020 & Wikiart & \textbf{Accuracy} 58.9 \cite{zhong2020fine} \\ \hline
\end{tabular}
\caption{Classification tasks on artworks and their related performance year by year.}
\label{tab:arttaskperf}
\end{table}

%The authors of \cite{shen2019discovering} focus their research on finding near duplicate patterns in collections of artworks and introduce a novel dataset ... They show that ....
 
 %\cite{yin2016object},  . 

%\textcolor{red}{LA PARTE SUI DATA SET  VA VERIFICATA E AGGIORNATA  MOLTO BENE, OMNIART DI CERTO SUPPORTA STUDI ICONOGRAFICI}

\subsubsection{Artwork analysis data sets}
Machine learning methods are data-driven and thus the progress of research is conditioned by the availability of data sets for training algorithms. In the field of  art work 
image analysis great efforts have been devoted to the creation of data sets and baselines   \cite{wilber2017bam,mao2017deepart,strezoski2017omniart,gonthier2018weakly,bianco2019multitask,stefanini2019artpedia,strezoski2020omnieyes}. 
Table \ref{tab:artdatasets} surveys the data sets mentioned in the art work image analysis  literature. We can observe that the number of images varies between $\sim$ 4k and $\sim$ 2M, and the proposed annotations almost always include artist, style, genre, type, medium, and material. Some notable data sets are \textbf{Iconart}, which focuses on Christian art, \textbf{OmniArt}, which includes also IconClass categories, and \textbf{Painting-91}, which contains 10 classes belonging to the Pascal VOC data set. At present,  most art data sets offer whole-image labeling, and  only a few data sets provide the location of  objects within the image,  but only in small numbers and for a limited set of rather generic objects. The overview of Table \ref{tab:artdatasets} highlights the opportunity of producing more data sets with finer-grain annotations at the object and pixel level.

\begin{table}[t]
\centering
\small
\begin{tabularx}{\linewidth}{|X|p{2.7cm}|c|m{3cm}|m{2.5cm}|}
\hline
\textbf{Name} & \textbf{Domain} & \textbf{\#Images} & \textbf{Classes} & \textbf{Annotations} \\
\hline
\textbf{Art500k}  \cite{mao2017deepart, mao2019visual} & paintings generic & 550k & artist, genre, \newline art movement, event, historical figure, \newline description & image-level  \\
    \hline
\textbf{AVA Style}  \cite{karayev2013recognizing} & paintings generic & 14k & 14 styles & image-level \\
    \hline
\textbf{Flickr Style} \cite{karayev2013recognizing} & paintings generic & 80k & 20 styles & image-level \\
    \hline
\textbf{Iconart} \cite{gonthier2018weakly} & christian paintings  & 6k & 7 classes: 4 saints, \newline 3 generics & image-level, \newline bounding boxes \newline (only test set) \\
    \hline 
\textbf{Multitask Painting 100k} \cite{Bianco2019MultitaskPC} & paintings generic & 100k & 1508 artists, \newline 125 styles, \newline 41 genres & image-level  \\
    \hline
\textbf{OmniArt} \cite{strezoski2017omniart, strezoski2018omniart} & artworks generic & 2M & artist, collection, \newline type, school, century, source, region, \newline iconclasses, genre & image-level, \newline bounding boxes  \\
    \hline
\textbf{Painting-91} \cite{painting91} & paintings generic & 4k & 91 artists, \newline 13 styles & image-level \\
    \hline
\textbf{Paintings Dataset} \cite{BMVC.28.38, crowley2016art} & paintings generic & 8.6k & 10 VOC12 classes & image-level  \\
    \hline
\textbf{Pandora} \cite{florea2016pandora} & paintings generic & 8k & 12 styles & image-level \\ \hline
\textbf{PRINTART} \cite{carneiro2012artistic} & art generic & 1k & 75 classes & image-level, \newline bounding-boxes \\ \hline
\textbf{Rijkmuseum} \cite{MensinkICMIR2014, sabatelli2018deep} & paintings generic & 112k & 206 materials, \newline 1k types, \newline 1k artists & image-level  \\
    \hline
\textbf{Wikiart} \cite{karayev2013recognizing} & paintings generic & 85k & 25 styles & image-level \\
    \hline
\end{tabularx}
\caption{State-of-the-art data sets for classification on the artworks field with their corresponding number of images, classes and type of annotation.}
\label{tab:artdatasets}
\end{table}

%jocch
%\cite{funkhouser2011learning}

%- search pattern \cite{ubeda2019pattern}

%- Image restoration \cite{yeh2017semantic} #not in art

%- other applications \cite{afreen2018semantic} \cite{yang2019classification}

\mycomment{

\subsection{Weakly Supervised Learning}

Large scale labeled data availability has played a fundamental role towards the success of supervised ML algorithms \cite{krizhevsky2012imagenet} and, indeed, most DL state-of-the-art models \cite{krizhevsky2012imagenet,he2016deep} are typically trained with massive amounts of high quality fine-grained labeled data \cite{deng2009imagenet,lin2014microsoft}.
Nonetheless, such required data are usually difficult to gather when facing a new real-world problem, given that its manual annotation is a labor-intensive, tedious and expensive task, which may even require some experience or domain-specific knowledge \cite{mensink2014rijksmuseum}.

The underlying efforts, costs and times promote the exploration of alternative solutions to exploit cheaper sources of labeled data, e.g., using publicly available data sets \cite{deng2009imagenet} or limited, partially, noisy and weakly annotated data, which can be easily obtained from the Web. \cite{sukhbaatar2014learning} proposed multiple approaches to train DL models with high volumes of noisy data, achieving competitive results w.r.t. models trained with larger amounts of high-quality  data.
Effective techniques for limited training data availability include:  transfer learning \cite{pan2009survey, lu2020knowledge}, to exploit common knowledge obtained from models pre-trained on a different domain or task; data augmentation \cite{perez2017effectiveness}, to generate new images through heuristic transformations over original labeled images (e.g. rotation, flip, crop, zoom, etc);  synthetic data generation with Generative Adversarial Networks (GANS); semi-supervised learning \cite{chapelle2009semi, tang2017visual}, to take advantage of both labeled and unlabeled data; and weakly supervised learning approaches \cite{zhou2018weakly}, to exploit labels with low-quality or higher abstraction relevance (e.g., use of ground truth bounding boxes  to perform a pixel-level segmentation task).
All such general purpose techniques can be combined in order to maximize the performance of hybrid methods.

Finally, the application of weakly supervised learning for art studies is of particular interest for the research described in this paper. \cite{crowley2013gods} studied a weakly supervised approach to identify figurative art in decorations of Greek vases by exploiting a data set of images with associated brief text descriptions. \cite{gonthier2018weakly} presented a data set for weakly supervised object detection in paintings and proposed an approach to find iconographic elements based on transfer learning. The authors used a pre-trained Faster R-CNN model \cite{girshick2015fast}, to detect candidate bounding boxes and extract corresponding features, and assigned labels and confidence by means of a simple multiple-instance learning method.
}

\subsection{Interpretability and attention maps}

Deep Learning architectures are  black-boxes with extremely good performance but hard to analyze in their internal mechanics.  The opaqueness of their behavior makes the diagnosis of errors and the optimization of inference problematic. Various efforts have been made to design intuitive  techniques for the visualization of CNN behavior supporting the  interpretation of their performance. A prominent result in this direction is the concept of Class Activation Map (CAM). CAMs,  first introduced in \cite{zhou2016learning},  aim at highlighting for a specific object class the most discriminative areas of the input image that contribute to the assignment to that class. The idea is to map the  class weights at the output of the CNN back to features maps of the last convolutional layer from which they are computed using global average pooling. By up-sampling such a projection to the size of the input image, one can visualize the importance of each image location for the prediction of the class.  
The original  formulation of the CAM has been subsequently extended. The work in \cite{selvaraju2017grad} introduces Gradient-weighted Class Activation Mapping (Grad-CAM) and exploits the gradients of any target object through the final convolution layer to produce a map of the areas most contributing to the activation. Such an  approach  can be applied to other tasks beyond classification and does not require changing or re-training of the architecture. The Guided Grad-CAM  method  further combines low-resolution class-discriminative heat maps from Grad-CAM and fine-grained details extracted with guided back-propagation \cite{springenberg2014striving}.  \cite{chattopadhay2018grad} proposes Grad-CAM++, a method that differentiates the importance of each pixel in a feature map yielding to an improvement with respect to Grad-CAM in the localization of single and multiple instances. The authors of \cite{omeiza2019smooth} modify the approach of \cite{chattopadhay2018grad} to obtain Smooth Grad-Cam++. As the name suggests, a smoothing   of the gradients is introduced  by adding small perturbations to the image of interest and making an average of all gradient matrices generated from the noisy images. Their method can also   create visualizations for specific layers, feature maps or neurons. %\cite{morbidelli2020augmented}

In Section \ref{sec:qualitative} we exploit the CAMs to investigate how the proposed classifier identifies the designated iconography classes in the images. 
\section{A Deep Learning Method for Iconography Classification}\label{sec:method}

In this paper, we present a CNN model for the task of  identifying   saint iconography in Christian art paintings. The model is  trained and tested on the ArtDL data set, which we have created on purpose for the task and made public at the address \texttt{https://www.artdl.org}.   

\subsection{Data set}\label{sec:meth-dataset}
The ArtDL data set contains 42,479 images of artworks portraying Christian saints, divided in 10 classes: Saint Dominic (iconclass 11HH(DOMINIC)), Saint Francis of Assisi (iconclass 11H(FRANCIS)), Saint Jerome (iconclass 11H(JEROME)), Saint John the Baptist (iconclass 11H(JOHN THE BAPTIST)), Saint Anthony of Padua (iconclass 11H(ANTONY OF PADUA), Saint Mary Magdalene (iconclass 11HH(MARY MAGDALENE)), Saint Paul (iconclass 11H(PAUL)), Saint Peter (iconclass 11H(PETER)), Saint Sebastian (iconclass 11H(SEBASTIAN)) and Virgin Mary (iconclass 11F). All images are associated with high-level annotations specifying which iconography classes appear in them (from a minimum of 1 class to a maximum of 7 classes).

\begin{table}[t]
\centering
%\begin{tabular}{ | >{\centering\arraybackslash}m{6cm} | >{\centering\arraybackslash}m{4cm} | } 
\begin{tabular}{ | p{6cm} | c | } 
\hline
\textbf{Source} & \textbf{\#Images}\\ 
\hline
Catalogo Generale dei Beni Culturali\cite{benicult} & 21,749\\ \hline
Gallerix\cite{gallerix} & 3,110\\ \hline
Museo Nacional del Prado\cite{prado} & 1,285\\ \hline
National Gallery of Art\cite{nga} & 289\\ \hline
Pharos\cite{pharos} & 1,938\\ \hline
Rijksmuseum\cite{rijkmus} & 471\\ \hline
The Metropolitan Museum of Art\cite{met} & 675\\ \hline
Web Gallery of Art\cite{wga} & 6,740\\ \hline
Wikiart\cite{wikiart} & 7,646\\ \hline
Wikidata\cite{wikidata} & 3,648\\ \hline
\textbf{Total} & \textbf{47,551}\\ \hline
\end{tabular}
\caption{Amount of images retrieved from each source}
\label{tab:nsources}
\end{table}

\subsubsection{Image acquisition}

\begin{figure}[t]
    \centering
    \includegraphics[width=0.8\linewidth]{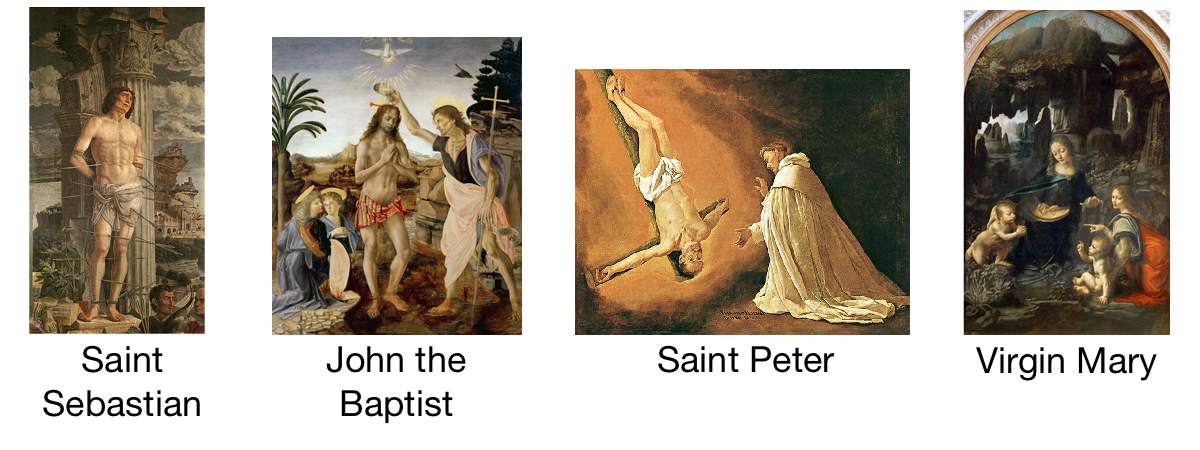}
    \caption{Examples of images in the data set and their annotations.}
    \label{figure:sampleimages}
\end{figure}

The total number of collected images before any filtering is 47,551, gathered from the data sources listed in  Table \ref{tab:nsources}. Images were retrieved by using public APIs, CSV databases or web scraping.   The data set comprises both RGB (60\%) and BW (40\%) images. Most   sources contained only RGB images while some sources, e.g., Catalogo Generale dei Beni Culturali \cite{benicult}, published almost only BW images.
All images were acquired together with the information about them available in the data source, including artwork title, description, tags and other metadata.
Figure \ref{figure:sampleimages} illustrates some examples of the collected images.

\subsubsection{Image filtering} After  automatic image acquisition, three main data quality issues had to be addressed:

\textbf{Duplicate images.} Some sources are themselves a collection of multiple open data sets and thus  exact and near-duplicates may occur.  The detection and removal of such images were performed in two steps: first by automatically removing all the files with the same MD5 hash and then by calculating the hash similarity of image pairs and submitting to the user those pairs with a similarity above a certain threshold.  

\textbf{Damaged, empty, and non-painting images.} Several images portray ruined  or incomplete artworks or were wrongly annotated with  classes not present in the artwork. To quickly filter as many irrelevant  images as possible we exploited a state-of-the-art model for face and pose detection  (OpenPose \cite{8765346}) and removed all the images with at least one annotation but no detected  pose. Furthermore, we manually inspected  and possibly included by manually annotating the correct classes those images with  one detected pose but no annotations.

\textbf{Fragment images.} Several  images do not represent an entire artwork but only some part of it, e.g. a single panel of a polyptych or a specific part of a scene. All the images with metadata  including  keywords  such as \quoted{detail}, \quoted{fragment}, or \quoted{portion}, were manually processed. Some of them were kept in the data set because they contained specific iconography hints  (e.g., an image containing only the ointment jar of \textit{Mary Magdalene}).

\subsubsection{Annotation creation and revision}

The data set was initially labelled with iconography class keywords found in the artwork title, description, tags and other metadata. The automatically assigned labels have been validated by manually inspecting a random sample of $\sim30\%$ of the items. In  $~5\%$ of the cases we found that  keywords were misleading or too broad (e.g. \textit{John} or \textit{Mary} without further specification) and referred to unrelated paintings. 

\begin{table}[t]
\centering
\begin{tabularx}{\linewidth}{|X|c|c|c|c|c|c|c|c|c|c|c|}
\hline
Class & Anth. P. & Francis A. & Jerome & John B. & Magd. & Paul & Peter & Dom. & Sebas. & Virg. & None \\ \hline
\# Train & 171 & 1203 & 1272 & 1249 & 1936 & 750 & 1459 & 385 & 620 & 15492 & 12620\\ \hline
\# Val & 21 & 151 & 159 & 156 & 242 & 94 & 183 & 48 & 78 & 1937 & 1562\\ \hline
\# Test & 22 & 150 & 159 & 156 & 242 & 94 & 182 & 48 & 77 & 1936 & 1568\\ \hline
\# Total & 214 & 1504 & 1590 & 1561 & 2420 & 938 & 1824 & 481 & 775 & 19365 & 15750 \\ \hline
\end{tabularx}%
\caption{Total number of images for each class. \textit{None} represents the images with no annotated class. The number of images used for training, validation and test is also reported.}
\label{tab:nsaints_table}
\end{table}

\subsubsection{Data set split}
The data set is split into training (33,983 images), validation (4,248 images) and test (4,248 images) by keeping  the distribution of each class in the 3 splits balanced, e.g. 80\% of Virgin Mary in the training, 10\% in the validation and 10\% in the test set. Table \ref{tab:nsaints_table} shows the total number of images for each class and the number of images with no annotated class. There is a sensible imbalance among the classes, with \textit{Virgin Mary} being depicted in 19,365 images while \textit{Anthony of Padua} in 214 images.  

\begin{comment}
%tabella con co-occurrences
\begin{table}[ht]
\centering
\begin{tabularx}{\linewidth}{|c|X|X|X|X|X|X|X|X|X|X|}
\hline
 & Cath. & Franc. & Jer. & John & Jos. & Magdal. & Paul & Peter & Seb. & Mary \\ \hline
Cath. & 1437 & 44 & 34 & 72 & 98 & 80 & 32 & 55 & 26 & 528 \\ \hline
Franc. & 44 & 1576 & 66 & 72 & 75 & 54 & 32 & 50 & 17 & 446 \\ \hline
Jer. & 34 & 66 & 1590 & 70 & 35 & 42 & 30 & 40 & 16 & 347 \\ \hline
John & 72 & 72 & 70 & 1561 & 54 & 75 & 66 & 103 & 31 & 499 \\ \hline
Jos. & 98 & 75 & 35 & 54 & 5955 & 145 & 32 & 54 & 11 & 4816 \\ \hline
Magdal. & 80 & 54 & 42 & 75 & 145 & 2420 & 29 & 60 & 17 & 1578 \\ \hline
Paul & 32 & 32 & 30 & 66 & 32 & 29 & 1171 & 355 & 8 & 241 \\ \hline
Peter & 55 & 50 & 40 & 103 & 54 & 60 & 355 & 2204 & 18 & 425 \\ \hline
Seb. & 26 & 17 & 16 & 31 & 11 & 17 & 8 & 18 & 775 & 166 \\ \hline
Mary & 528 & 446 & 347 & 499 & 4816 & 1578 & 241 & 425 & 166 & 20687 \\ \hline
\end{tabularx}%
\caption{Co-occurrence matrix of the 10 classes of ArtDL.}
\label{tab:cooc_table}
\end{table}
\end{comment}

\begin{figure}[t]
    \centering
    \includegraphics[width=0.6\linewidth]{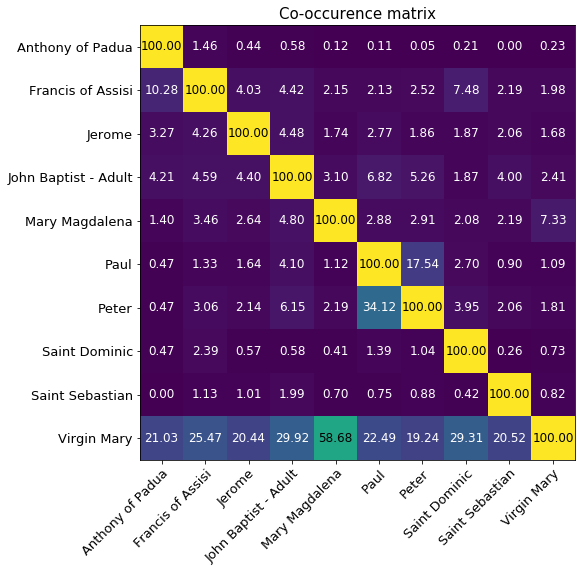}
    \caption{Co-occurrence matrix of the 10 classes present in the training split of ArtDL. Each value is the percentage of images of class X that are belonging also to class Y, e.g. 21.03\% images labelled with \textit{Anthony of Padua} are annotated with \textit{Virgin Mary} too.
    }
    \label{figure:coocc}
\end{figure}

Figure \ref{figure:coocc} shows the distribution and the co-occurrence of classes in the training set. Co-occurrence  gives the percentage of images of class X that contain also an annotation for class Y, e.g. $58.68\%$ of images annotated with \textit{Mary Magdalene} are annotated also with \textit{Virgin Mary}. The validation and test set have a similar distribution. Some classes have a high co-occurrence (e.g., \textit{Peter} and \textit{Paul} or \textit{Mary Magdalene} and \textit{Virgin Mary}). 

%\newpage %new page per risolvere un bug in cui il testo continuava anche dopo la fine della pagina
High co-occurrence values are due to the presence in the data set of polyptychs and of  complex scenes depicting multiple characters. Such frequently co-occurring classes may create confusion in the model and be more difficult to distinguish.

\subsection{Architecture}\label{sec:archres}

The developed classifier exploits a backbone based on ResNet50 \cite{he2016deep}  trained on the ImageNet data set \cite{deng2009imagenet}.
Such architecture is made fully convolutional by replacing the last two layers (average pooling and fully connected layers) with a 1x1 convolution layer, which outputs one channel for each class of the data set and acts as a classifier. We decided to use a pre-trained model instead of performing the training from scratch, due to the limited number of training images available, their high complexity and variability (many of them contain difficult scenes with numerous subjects) and the complexity of the network. %\textcolor{red}{verificare che non sia possibile creare (o gia disponibile) un modello addestrato su omniart che ha 2M di immagini}

During the training phase, images are transformed and augmented. First of all, they are padded to a square, based on their largest dimension, then they are resized to a fixed square size and normalized by the mean and standard deviation of the data set. The padding step is useful to avoid distortions and keep the aspect ratio unmodified after the resizing, while the resizing is useful to train on batches of multiple samples and to fit more images in the GPU memory. The augmentation performed is an horizontal flip with a probability of 50\%.

\subsection{Fine-tuning}\label{sec:meth-fine-tuning}
Transfer Learning (TL) is an effective technique that exploits the feature extraction capability of a model trained in a given domain to support a task in a different one. TL works by fine tuning, i.e., by retraining only selected  layers of the original model, normally the deepest ones that work at an abstraction level that is more domain-dependent. TL has been  recently shown to improve performance in art-related classification and detection tasks \cite{westlake2016detecting,cetinic2018fine}.  We apply TL to the original ResNet50 architecture and fine-tune all the layers except the initial 7x7 convolution, the 3x3 max pooling layers, and the first two residual blocks.
%LOW VS TOP LAYERS FOR FINE TUNING: We retrain both the top and the bottom layer of the network, responsible for high-level classification output and low level features detection respectively. By transforming natural images intro drawings. The different nature of the classes of objects requires re-training the top layers of the CNN model, because the fine structures and features of drawings are different as well, this motivate us to also re-train the bottom layers of the CNN
%reference: Object recognition in art drawings: transfer of a neural network, maybe to take into account for future experiments,  
This allows the model to exploit both low-level features, which are more general purpose (e.g. edges, corners, shapes, etc.), learned from a great amount of natural images, and mid and high-level features learned from artworks during fine-tuning. The fine-tuned layers were trained with a lower learning rate with respect to the 1x1 convolution layer, to gradually adapt the previously learned features to the artwork domain.

\subsection{Output}
The outputs of the network are:
\begin{itemize}
    \item A value for each class denoting  the confidence of the classification.
    \item Class-aware cues, represented by the Class Activation Maps (CAMs) \cite{zhou2016learning}, which highlight the areas of the image contributing most to its  assignment to a class.
\end{itemize}

\section{Results}\label{sec:results}

In this section we present a quantitative and qualitative analysis of the iconography classification performances on the ArtDL data set. Experiments were performed on a subset composed of  single-label images of the 10 selected classes. Note that being single-label does not imply that the image contains only one character. A complex scene can be annotated  with only  one class and is still considered single-label, as can be seen in Figure \ref{figure:fp}. Exploiting images without annotations and with multiple labels  is part of our future work.

The image subset used in the experiments contains 18,637 images: 14,912 for training, 1,861 for validation and 1,864 for testing.

\begin{figure}[t]
    \centering
    \includegraphics[width=0.7\linewidth]{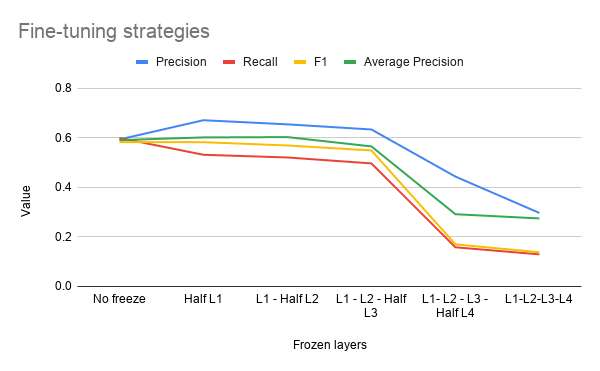}
    \caption{Performances of the model when freezing at a specified layer and fine-tuning the following deeper layers. For each metrics, a single value is obtained by averaging across all the  classes. }%\textcolor{red}{Immagine da aggiornare con nuovi dati}
    \label{figure:finetuning}
\end{figure}

\subsection{Quantitative results} \label{sec:quantitative}
The evaluation metrics used for the quantitative analysis are: Precision, Recall, Average Precision (AP) and Confusion Matrix. Due to the high imbalance of the data set, Accuracy is not  appropriate to assess the performances of the model.

\subsubsection{Fine-tuning} 
As explained in Section \ref{sec:meth-fine-tuning}, the architecture used in this work is a Fully Convolutional ResNet50 network pre-trained on ImageNet. A comparison of alternative configurations has been performed to determine the most effective freezing strategy to apply during fine tuning. Figure \ref{figure:finetuning} shows the performances obtained when freezing the architecture at different layers. The best results are obtained by freezing the first few layers, which learn low-level features, and by fine-tuning all the deeper layers;  training  only the classifier while freezing all the pre-trained layers yielded the worst results. These findings are consistent with the literature \cite{yosinski2014transferable}: fine-tuning almost all the layers of the architecture is required due to the different domain and nature of ImageNet (the data set used to pre-train the architecture) and the targeted iconography data set.

\subsubsection{Data set resampling}
As mentioned in  Section \ref{sec:meth-dataset} the ArtDL data set is characterized by a rather high class imbalance. Several techniques can be exploited to address class imbalance during training. We compared  the techniques of weighted loss,  undersampling of the majority classes, and oversampling of the minority classes. Oversampling of the minority classes to the majority class \textit{Virgin Mary}, applied before training. proved the best option and yielded a sensible improvement to both quantitative and qualitative results.

\begin{table}[t]
\begin{small}
\centering
\resizebox{\textwidth}{!}{%
\begin{tabular}{|c|c|c|c|c|c|}
\hline
\textbf{Class name} & \textbf{\# Test Images} & \textbf{Precision} & \textbf{Recall} & \textbf{F1-Score} & \textbf{Average Precision} \\ \hline
Anthony of Padua & 14 & 72.73\% & 57.14\% & 64.00\% & 64.14\%\\ \hline
Francis of Assisi & 98 & 69.23\% & 82.65\% & 75.35\% & 76.06\%\\ \hline
Jerome & 118 & 70.77\% & 77.97\% & 74.19\% & 78.88\%\\ \hline
John the Baptist & 99 & 58.09\% & 79.80\% & 67.23\% & 75.69\%\\ \hline
Mary Magdalene & 90 & 79.27\% & 72.22\% & 75.58\% & 82.23\%\\ \hline
Paul & 52 & 54.55\% & 34.62\% & 42.35\% & 38.47\%\\ \hline
Peter & 119 & 72.95\% & 74.79\% & 73.86\% & 77.93\%\\ \hline
Saint Dominic & 29 & 50.00\% & 65.52\% & 56.72\% & 54.35\%\\ \hline
Saint Sebastian & 56 & 91.11\% & 73.21\% & 81.19\% & 82.46\%\\ \hline
Virgin Mary & 1189 & 93.04\% & 91.00\% & 92.01\% & 97.03\%\\ \hline
\multicolumn{2}{|c|}{\textbf{Mean}} & 71.17\% & 70.89\% & 70.25\% & 72.73\% \\ \hline
\end{tabular}%
}
\caption{Evaluation metrics computed on the test set. For each class we report the number of test images and the values of the metrics used to evaluate the performances of the model. The last row contains the mean value over all the classes.}
\label{tab:prec_rec}
\end{small}
\end{table}

\subsubsection{Classification evaluation}
Table \ref{tab:prec_rec} shows the evaluation results on the test set for each iconography class. Almost all the classes have more than 68\% of precision with some reaching $\sim85$\% while having $\sim60$\% of recall.  \textit{Virgin Mary} is the class with the best performances, as expected due to  the high number of images belonging to this class. \textit{Paul} is the one with the worst performances: this result can be attributed to its rather generic iconography: the saint is most often represented as a common bearded man and only in very few images has the sword in his hand, which is the distinctive attribute of the class. Other classes, such as \textit{Saint Sebastian}, have a number of images  similar to \textit{Paul} but their distinctive iconography  attributes are much more  frequently displayed in the artworks. The obtained results are consistent and even better with respect to those found in the previous research for artwork content analysis (reported also in Table \ref{tab:arttaskperf}). In \cite{gonthier2018weakly} and \cite{gonthier2020multiple}, which are the works most closely related to ours, the authors  respectively report a mean AP of $62.7\%$ and $69.2\%$ for the classification task; \cite{crowley2016art} reaches a mean AP of $68.5\%$ on the classification of generic objects in artworks. These figures are obtained with  different  tasks and data sets and therefore do not constitute a head-to-head comparison.

\begin{figure}[t]
    \centering
    \includegraphics[width=0.7\linewidth]{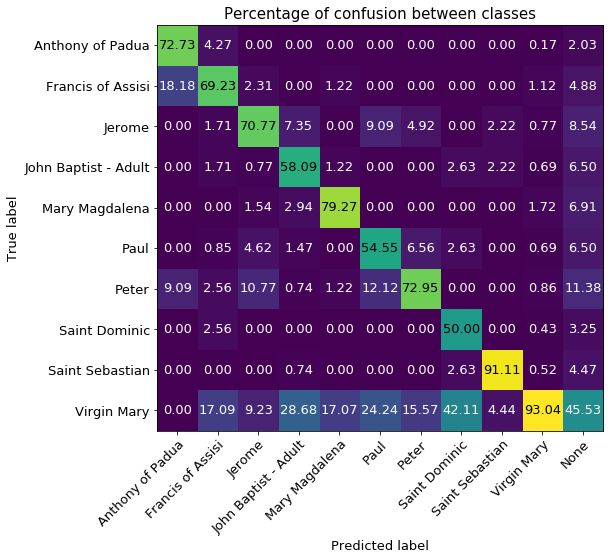}
    \caption{Confusion between classes. Rows are the ground truth classes and columns are the predicted classes. Values on the main diagonal are the precision of each class and values on the columns  add up to 1.
    }
    \label{figure:conf_classes}
\end{figure}

Figure \ref{figure:conf_classes} shows the percentage of confusion between classes. Such an analysis allows one to better understand which class Y is predicted instead of the real class X and how frequent is such a wrong prediction, easing the diagnosis of classification errors.
For completeness, the table includes also the \textit{None} label, to address the images with a ground truth class but no predicted classes. The table highlights the quite strong discriminative power of the model which   distinguishes well most classes. Some  cases deserve attention:    \textit{Saint Dominic} is wrongly predicted $\sim42$\% of the times as \textit{Virgin Mary}. This is due to the fact that the respective images are the least present in the data set  and the model falls back to the most represented class. Adding more and diverse examples of the class would eliminate the confusion.

\subsection{Qualitative results} \label{sec:qualitative}

The iconography of Christian saints relies on the presence of specific symbolic attributes \cite{Lanzi04}. This makes the classification task particularly interesting, because one can use the interpretation tools described in Section \ref{sec:relwork} to understand how the model mimics the human behavior in recognizing a saint, either by looking at the distinctive attributes or at the global context in which the character is embedded. For this purpose, we examined the CAMs produced by our architecture for a number of sample images.

\begin{figure}[ht]
\includegraphics[width=1\linewidth]{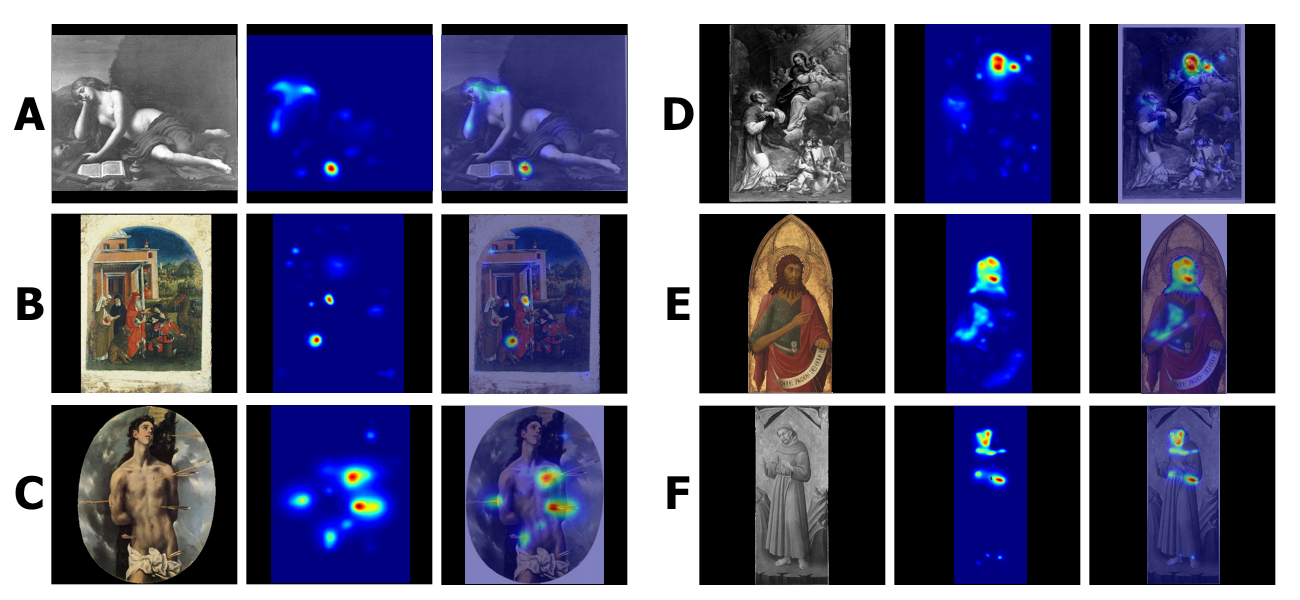}
\centering
\caption{Examples of Class Activation Maps on different test images and for different classes. (A) St. Mary Magdalene, (B) St. Jerome, (C) St. Sebastian, (D) Virgin Mary, (E) St. John the Baptist, (F) St. Francis of Assisi.}
\label{figure:good_cams}
\end{figure}

Figure \ref{figure:good_cams} contains six test images, depicting different scenes, in which at least an example for each class is visible. For each original test image (left), we also show the CAM (center) \textit{of the most prominent class for each example} and the original image overlaid with the CAM (right). 

Figure \ref{figure:good_cams}.A\footnote{"Maddalena penitente", Turchi Alessandro detto Orbetto, 1635-1640} represents \textit{St. Mary Magdalene}. It is noticeable that the area that activates the network  the most is the ointment jar in the bottom center of the painting,  one of the identifying symbols of Mary Magdalene. We can also note  a less strong activation induced by the presence of her hairs, which are a well-known characterizing attribute of the saint. In other paintings this symbol yields a very high activation.

Figure \ref{figure:good_cams}.B\footnote{"I mercanti che avevano rubato l'asino chiedono perdono al santo", Maestro dei Gesuati, 1450-1459} depicts \textit{St. Jerome} in the desert. From the CAM, we can see that the model learns to recognize the saint by his face and beard. As for St. Magdalene, the model is also able to capture a specific symbol used for identifying St. Jerome, the lion; which has a very high activation. In other paintings, the area with the highest activation is his red flat-top cardinal's hat.

If we look at Figure \ref{figure:good_cams}.C\footnote{"St. Sebastian", El Greco, 1600}, we can see \textit{St. Sebastian} who is recognizable by his body tied to a tree and shot with arrows. The CAM shows that the arrows correspond to the areas that contribute most to the classification and are extracted as the salient features of the recognized subject  following the iconography very precisely. We can observe that the naked torso of \textit{St. Sebastian} is not useful to recognize the class because there are a lot of scenes with other naked men, e.g., \quoted{The Crucifixion of Jesus} or \quoted{The Baptism of Jesus}.

Figure \ref{figure:good_cams}.D\footnote{"Apparizione della Madonna a San Filippo", Ricci Ubaldo, 1700-1749} shows the \textit{Virgin Mary} holding the Child Jesus. Here, we can see that the main characteristic that the model uses to classify Mary is her veiled head. But the activation  area covering the Child Jesus shows the utility of the context for the recognition;  the model has learned that also Child Jesus is important to recognize \textit{Mary}. In the iconography and in the training images, these two subjects are in most cases very close to each other and sometimes even overlapping.

Figure \ref{figure:good_cams}.E\footnote{"Saint John the Baptist", Lippo Memmi, 1330 - 1340} shows an example where the model correctly classifies \textit{St. John the Baptist}. The most activating region is the head of the saint; another area useful for the classification task is the saint's ragged clothing. Saint John the Baptist  usually appears in complex scenes with other male characters: by looking at the CAMs in other paintings, e.g., representations of  "The baptism of Christ", one can note that the saint is often confused with Jesus Christ or the naked body of other men.

Figure \ref{figure:good_cams}.F\footnote{"San Francesco", Domenico di Michelino, 1460} shows \textit{Francis of Assisi}. As we can see, the saint is mainly recognized by his face and  by the white cord of his clothing, which is a specific iconography symbol for this class. Even if the training classes include \textit{Anthony of Padua}, who is represented very similarly to \textit{Francis of Assisi}, the classifier differentiates  the two classes well (Figure \ref{figure:conf_classes}).

From the presented examples and from many other similar cases, we can observe that the proposed architecture, even if based on a quite simple architecture exploiting a model pre-trained   on natural images and fine-tuned with rather noisy artwork images, learns to discriminate well the  iconography classes using the same clues that we, as humans, consider important when describing the themes of a painting.

%Some of the features, for example the face,  upper torso, and the arms of the crucified Jesus are  exactly the features on which artists have expressed the most interesting and historically and culturally relevant variants of the iconography \cite{vavala29}.

\begin{figure}[t]
    \centering
    \includegraphics[width=0.8\linewidth]{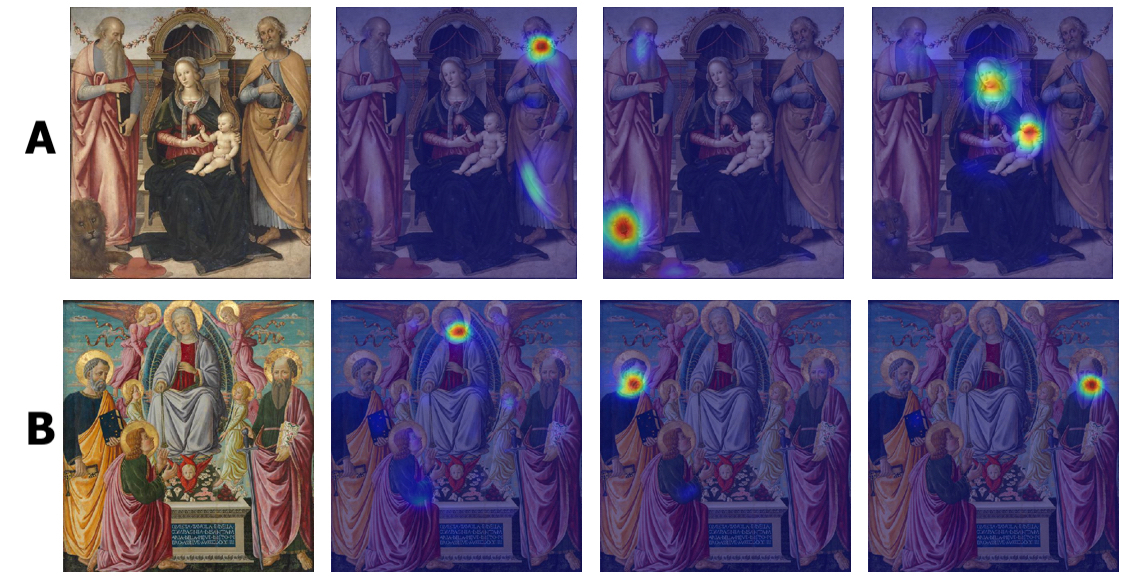}
    \caption{Examples of images with missing ground truth annotations that the model is able to correctly extract. For each image here we present: the original painting and the original painting overlaid  with the CAMs of the depicted classes.
    }
    \label{figure:fp}
\end{figure}

\subsubsection{False Positive  analysis}

Mistaken FP predictions can happen because the model correctly predicts the classes portrayed in the image but the ground truth annotations are wrong or missing. This motivation is especially relevant in our case in which annotations were obtained semi-automatically and are both scarce and noisy. 

Figure \ref{figure:fp} shows two cases of false false positives, i.e., images for which the model correctly predicts the classes, but the noisy ground truth misses the annotations and thus the predictions are mistakenly counted as a false positive. Figure \ref{figure:fp}.A\footnote{"La vergine in trono con il Bambino e tra san Girolamo e san Pietro", Andrea d'Assisi, 1490} shows \textit{Virgin Mary} holding Child Jesus between \textit{Jerome} and \textit{Peter}. The model classifies with over 90\% probability the presence of \textit{Jerome} and \textit{Virgin Mary} even if only \textit{Peter} is included in the annotations of the image. The reported CAMs show the attention area. These results are consistent with the test images presented in Figure \ref{figure:good_cams}: \textit{Jerome} is recognized by the lion at his feet and \textit{Virgin Mary} is localized by her face and the contextual presence of the Child.

Figure \ref{figure:fp}.B\footnote{"Madonna della Cintola", Bernardo di Stefano Rosselli, 1480} shows \textit{Virgin Mary} holding Child Jesus and surrounded by  saints. In this case, \textit{Virgin Mary} is the only annotated class but the model can also correctly recognize \textit{Peter} and \textit{Paul}. From the CAMs, we can see that there is no  confusion between the three classes.

These results suggest that we could use the output of the model, especially when there is a strong classification score, to iteratively refine the data set labels and thus increase the data set completeness and the model performance. The same approach could be used to support the extension of the data set, by automatically proposing candidate annotations for  the  new images.

\section{Conclusions and Future Work}\label{sec:concl}

In this work, we have presented a novel artworks iconography data set (ArtDL) and a simple architecture  for the classification of iconographic entities in Christian art paintings. Evaluation results show good performance even in presence of the label noise produced by the semi-automatic procedure employed for data collection.  

Future work will concentrate on the following  objectives;
\begin{itemize}
    \item Data set improvement and extension: we plan to use the current classifier to reduce the label noise and to incorporate more iconography classes. 
    \item Architecture improvement: we aim at improving the classifier by applying techniques for learning with scarce labels (e.g., label refinery \cite{labelrefinery}) and semi-supervised learning methods (e.g., classifier improvement with Generative Adversarial Networks \cite{Salimans2016}). The resulting model will be also applied to multi-object classification.
    \item Object detection and instance segmentation support: iconography studies require not only the detection of classes but also of their position and mutual spatial relations. To this end, we are building an object detection and instance segmentation architecture, exploiting the CAMs of the classifier to enable the semi-automatic creation of the bounding boxes and pixel-level annotations required for training object detectors, along the lines of \cite{gonthier2018weakly} and \cite{ahn2019weakly}. 
\end{itemize}

Our ultimate goal is to cover the entire IconClass dictionary, which features 28k classes. This long term goal requires effective IT-enabled tools for the annotation process, which can be supported by the automatic classification of single characters and by the  inference of more complex classes  comprising multiple individuals in specific spatial relations (e.g., by predicting the class \quoted{Ann seldbritt}  (IconClass 73A221) when Saint Ann appears holding  the Virgin Mary and the baby Jesus in her lap). The envisioned iconography classifier and localization detector could support    the study of the evolution of iconography across space and time, enabling advanced queries over art images collections that mix visual content and metadata (e.g., \quoted{Find all annunciations where the angel appears to the right of Mary} or \quoted{Find the earliest  three-nails crucifixion in the collection}).

\bibliographystyle{ACM-Reference-Format}
\bibliography{bibi}


\begin{thebibliography}{90}

%%% the Web address.

\ifx \showCODEN    \undefined \def \showCODEN     #1{\unskip}     \fi
\ifx \showDOI      \undefined \def \showDOI       #1{#1}\fi
\ifx \showISBNx    \undefined \def \showISBNx     #1{\unskip}     \fi
\ifx \showISBNxiii \undefined \def \showISBNxiii  #1{\unskip}     \fi
\ifx \showISSN     \undefined \def \showISSN      #1{\unskip}     \fi
\ifx \showLCCN     \undefined \def \showLCCN      #1{\unskip}     \fi
\ifx \shownote     \undefined \def \shownote      #1{#1}          \fi
\ifx \showarticletitle \undefined \def \showarticletitle #1{#1}   \fi
\ifx \showURL      \undefined \def \showURL       {\relax}        \fi
% invisible to TeX
\providecommand\bibfield[2]{#2}
\providecommand\bibinfo[2]{#2}
\providecommand\natexlab[1]{#1}
\providecommand\showeprint[2][]{arXiv:#2}

\bibitem[\protect\citeauthoryear{??}{gal}{[n.d.]}]%
        {gallerix}
 \bibinfo{year}{[n.d.]}\natexlab{}.
\newblock \bibinfo{booktitle}{\emph{Gallerix online museum}}.
\newblock
\urldef\tempurl%
\url{https://gallerix.org}
\showURL{%
\tempurl}


\bibitem[\protect\citeauthoryear{??}{ben}{[n.d.]}]%
        {benicult}
 \bibinfo{year}{[n.d.]}\natexlab{}.
\newblock \bibinfo{booktitle}{\emph{ICCD - Istituto Centrale per il Catalogo e
  la Documentazione}}.
\newblock
\urldef\tempurl%
\url{http://www.iccd.beniculturali.it}
\showURL{%
\tempurl}


\bibitem[\protect\citeauthoryear{??}{met}{[n.d.]}]%
        {met}
 \bibinfo{year}{[n.d.]}\natexlab{}.
\newblock \bibinfo{booktitle}{\emph{The Metropolitan Museum of Art}}.
\newblock
\urldef\tempurl%
\url{https://www.metmuseum.org}
\showURL{%
\tempurl}


\bibitem[\protect\citeauthoryear{??}{pra}{[n.d.]}]%
        {prado}
 \bibinfo{year}{[n.d.]}\natexlab{}.
\newblock \bibinfo{booktitle}{\emph{Museo Nacional del Prado}}.
\newblock
\urldef\tempurl%
\url{https://www.museodelprado.es}
\showURL{%
\tempurl}


\bibitem[\protect\citeauthoryear{??}{nga}{[n.d.]}]%
        {nga}
 \bibinfo{year}{[n.d.]}\natexlab{}.
\newblock \bibinfo{booktitle}{\emph{The National Gallery, London}}.
\newblock
\urldef\tempurl%
\url{https://www.nationalgallery.org.uk}
\showURL{%
\tempurl}


\bibitem[\protect\citeauthoryear{??}{pha}{[n.d.]}]%
        {pharos}
 \bibinfo{year}{[n.d.]}\natexlab{}.
\newblock \bibinfo{booktitle}{\emph{PHAROS: The International Consortium of
  Photo Archives}}.
\newblock
\urldef\tempurl%
\url{http://pharosartresearch.org}
\showURL{%
\tempurl}


\bibitem[\protect\citeauthoryear{??}{rij}{[n.d.]}]%
        {rijkmus}
 \bibinfo{year}{[n.d.]}\natexlab{}.
\newblock \bibinfo{booktitle}{\emph{Rijksmuseum – The Museum of the
  Netherlands}}.
\newblock
\urldef\tempurl%
\url{https://www.rijksmuseum.nl/en}
\showURL{%
\tempurl}


\bibitem[\protect\citeauthoryear{??}{wga}{[n.d.]}]%
        {wga}
 \bibinfo{year}{[n.d.]}\natexlab{}.
\newblock \bibinfo{booktitle}{\emph{Web Gallery of Art, searchable fine arts
  database}}.
\newblock
\urldef\tempurl%
\url{https://www.wga.hu}
\showURL{%
\tempurl}


\bibitem[\protect\citeauthoryear{??}{wik}{[n.d.]a}]%
        {wikiart}
 \bibinfo{year}{[n.d.]}\natexlab{a}.
\newblock \bibinfo{booktitle}{\emph{Wikiart.org - Visual Art Encyclopedia}}.
\newblock
\urldef\tempurl%
\url{https://www.wikiart.org}
\showURL{%
\tempurl}


\bibitem[\protect\citeauthoryear{??}{wik}{[n.d.]b}]%
        {wikidata}
 \bibinfo{year}{[n.d.]}\natexlab{b}.
\newblock \bibinfo{booktitle}{\emph{Wikidata}}.
\newblock
\urldef\tempurl%
\url{https://www.wikidata.org/wiki/Wikidata:Main\_Page}
\showURL{%
\tempurl}


\bibitem[\protect\citeauthoryear{Ahn, Cho, and Kwak}{Ahn et~al\mbox{.}}{2019}]%
        {ahn2019weakly}
\bibfield{author}{\bibinfo{person}{Jiwoon Ahn}, \bibinfo{person}{Sunghyun Cho},
  {and} \bibinfo{person}{Suha Kwak}.} \bibinfo{year}{2019}\natexlab{}.
\newblock \showarticletitle{Weakly Supervised Learning of Instance Segmentation
  with Inter-pixel Relations}. In \bibinfo{booktitle}{\emph{Proceedings of the
  IEEE Conference on Computer Vision and Pattern Recognition}}.
  \bibinfo{pages}{2209--2218}.
\newblock


\bibitem[\protect\citeauthoryear{Alameda-Pineda, Ricci, Yan, and
  Sebe}{Alameda-Pineda et~al\mbox{.}}{2016}]%
        {Alameda-Pineda_2016_CVPR}
\bibfield{author}{\bibinfo{person}{Xavier Alameda-Pineda},
  \bibinfo{person}{Elisa Ricci}, \bibinfo{person}{Yan Yan}, {and}
  \bibinfo{person}{Nicu Sebe}.} \bibinfo{year}{2016}\natexlab{}.
\newblock \showarticletitle{Recognizing Emotions From Abstract Paintings Using
  Non-Linear Matrix Completion}. In \bibinfo{booktitle}{\emph{Proceedings of
  the IEEE Conference on Computer Vision and Pattern Recognition (CVPR)}}.
\newblock


\bibitem[\protect\citeauthoryear{Amirshahi, Hayn-Leichsenring, Denzler, and
  Redies}{Amirshahi et~al\mbox{.}}{2014}]%
        {amirshahi2014jenaesthetics}
\bibfield{author}{\bibinfo{person}{Seyed~Ali Amirshahi},
  \bibinfo{person}{Gregor~Uwe Hayn-Leichsenring}, \bibinfo{person}{Joachim
  Denzler}, {and} \bibinfo{person}{Christoph Redies}.}
  \bibinfo{year}{2014}\natexlab{}.
\newblock \showarticletitle{Jenaesthetics subjective dataset: analyzing
  paintings by subjective scores}. In \bibinfo{booktitle}{\emph{European
  Conference on Computer Vision}}. Springer, \bibinfo{pages}{3--19}.
\newblock


\bibitem[\protect\citeauthoryear{Anwer, Khan, van~de Weijer, and
  Laaksonen}{Anwer et~al\mbox{.}}{2016}]%
        {anwer2016combining}
\bibfield{author}{\bibinfo{person}{Rao~Muhammad Anwer},
  \bibinfo{person}{Fahad~Shahbaz Khan}, \bibinfo{person}{Joost van~de Weijer},
  {and} \bibinfo{person}{Jorma Laaksonen}.} \bibinfo{year}{2016}\natexlab{}.
\newblock \showarticletitle{Combining holistic and part-based deep
  representations for computational painting categorization}. In
  \bibinfo{booktitle}{\emph{Proceedings of the 2016 ACM on International
  Conference on Multimedia Retrieval}}. \bibinfo{pages}{339--342}.
\newblock


\bibitem[\protect\citeauthoryear{Bagherinezhad, Horton, Rastegari, and
  Farhadi}{Bagherinezhad et~al\mbox{.}}{2018}]%
        {labelrefinery}
\bibfield{author}{\bibinfo{person}{Hessam Bagherinezhad},
  \bibinfo{person}{Maxwell Horton}, \bibinfo{person}{Mohammad Rastegari}, {and}
  \bibinfo{person}{Ali Farhadi}.} \bibinfo{year}{2018}\natexlab{}.
\newblock \showarticletitle{Label Refinery: Improving ImageNet Classification
  through Label Progression}.
\newblock \bibinfo{journal}{\emph{CoRR}}  \bibinfo{volume}{abs/1805.02641}
  (\bibinfo{year}{2018}).
\newblock
\showeprint[arxiv]{1805.02641}
\urldef\tempurl%
\url{http://arxiv.org/abs/1805.02641}
\showURL{%
\tempurl}


\bibitem[\protect\citeauthoryear{Banerji and Sinha}{Banerji and Sinha}{2016}]%
        {banerji2016painting}
\bibfield{author}{\bibinfo{person}{Sugata Banerji} {and}
  \bibinfo{person}{Atreyee Sinha}.} \bibinfo{year}{2016}\natexlab{}.
\newblock \showarticletitle{Painting classification using a pre-trained
  convolutional neural network}. In \bibinfo{booktitle}{\emph{International
  Conference on Computer Vision, Graphics, and Image processing}}. Springer,
  \bibinfo{pages}{168--179}.
\newblock


\bibitem[\protect\citeauthoryear{Bar, Levy, and Wolf}{Bar
  et~al\mbox{.}}{2014}]%
        {bar2014classification}
\bibfield{author}{\bibinfo{person}{Yaniv Bar}, \bibinfo{person}{Noga Levy},
  {and} \bibinfo{person}{Lior Wolf}.} \bibinfo{year}{2014}\natexlab{}.
\newblock \showarticletitle{Classification of artistic styles using binarized
  features derived from a deep neural network}. In
  \bibinfo{booktitle}{\emph{European conference on computer vision}}. Springer,
  \bibinfo{pages}{71--84}.
\newblock


\bibitem[\protect\citeauthoryear{Belhi, Bouras, and Foufou}{Belhi
  et~al\mbox{.}}{2018}]%
        {belhi2018towards}
\bibfield{author}{\bibinfo{person}{Abdelhak Belhi}, \bibinfo{person}{Abdelaziz
  Bouras}, {and} \bibinfo{person}{Sebti Foufou}.}
  \bibinfo{year}{2018}\natexlab{}.
\newblock \showarticletitle{Towards a hierarchical multitask classification
  framework for cultural heritage}. In \bibinfo{booktitle}{\emph{2018 IEEE/ACS
  15th International Conference on Computer Systems and Applications
  (AICCSA)}}. IEEE, \bibinfo{pages}{1--7}.
\newblock


\bibitem[\protect\citeauthoryear{Bianco}{Bianco}{2017}]%
        {bianco2017large}
\bibfield{author}{\bibinfo{person}{Simone Bianco}.}
  \bibinfo{year}{2017}\natexlab{}.
\newblock \showarticletitle{Large age-gap face verification by feature
  injection in deep networks}.
\newblock \bibinfo{journal}{\emph{Pattern Recognition Letters}}
  \bibinfo{volume}{90} (\bibinfo{year}{2017}), \bibinfo{pages}{36--42}.
\newblock


\bibitem[\protect\citeauthoryear{Bianco, Mazzini, Napoletano, and
  Schettini}{Bianco et~al\mbox{.}}{2019a}]%
        {Bianco2019MultitaskPC}
\bibfield{author}{\bibinfo{person}{Simone Bianco}, \bibinfo{person}{Davide
  Mazzini}, \bibinfo{person}{Paolo Napoletano}, {and} \bibinfo{person}{Raimondo
  Schettini}.} \bibinfo{year}{2019}\natexlab{a}.
\newblock \showarticletitle{Multitask Painting Categorization by Deep
  Multibranch Neural Network}.
\newblock \bibinfo{journal}{\emph{Expert Syst. Appl.}}  \bibinfo{volume}{135}
  (\bibinfo{year}{2019}), \bibinfo{pages}{90--101}.
\newblock


\bibitem[\protect\citeauthoryear{Bianco, Mazzini, Napoletano, and
  Schettini}{Bianco et~al\mbox{.}}{2019b}]%
        {bianco2019multitask}
\bibfield{author}{\bibinfo{person}{Simone Bianco}, \bibinfo{person}{Davide
  Mazzini}, \bibinfo{person}{Paolo Napoletano}, {and} \bibinfo{person}{Raimondo
  Schettini}.} \bibinfo{year}{2019}\natexlab{b}.
\newblock \showarticletitle{Multitask Painting Categorization by Deep
  Multibranch Neural Network}.
\newblock \bibinfo{journal}{\emph{Expert Systems with Applications}}
  (\bibinfo{year}{2019}).
\newblock


\bibitem[\protect\citeauthoryear{Brachmann, Barth, and Redies}{Brachmann
  et~al\mbox{.}}{2017}]%
        {brachmann2017using}
\bibfield{author}{\bibinfo{person}{Anselm Brachmann}, \bibinfo{person}{Erhardt
  Barth}, {and} \bibinfo{person}{Christoph Redies}.}
  \bibinfo{year}{2017}\natexlab{}.
\newblock \showarticletitle{Using CNN features to better understand what makes
  visual artworks special}.
\newblock \bibinfo{journal}{\emph{Frontiers in psychology}}
  \bibinfo{volume}{8} (\bibinfo{year}{2017}), \bibinfo{pages}{830}.
\newblock


\bibitem[\protect\citeauthoryear{Brachmann and Redies}{Brachmann and
  Redies}{2017}]%
        {brachmann2017computational}
\bibfield{author}{\bibinfo{person}{Anselm Brachmann} {and}
  \bibinfo{person}{Christoph Redies}.} \bibinfo{year}{2017}\natexlab{}.
\newblock \showarticletitle{Computational and experimental approaches to visual
  aesthetics}.
\newblock \bibinfo{journal}{\emph{Frontiers in computational neuroscience}}
  \bibinfo{volume}{11} (\bibinfo{year}{2017}), \bibinfo{pages}{102}.
\newblock


\bibitem[\protect\citeauthoryear{Cai, Wu, Corradi, and Hall}{Cai
  et~al\mbox{.}}{2015}]%
        {cai2015cross}
\bibfield{author}{\bibinfo{person}{Hongping Cai}, \bibinfo{person}{Qi Wu},
  \bibinfo{person}{Tadeo Corradi}, {and} \bibinfo{person}{Peter Hall}.}
  \bibinfo{year}{2015}\natexlab{}.
\newblock \showarticletitle{The cross-depiction problem: Computer vision
  algorithms for recognising objects in artwork and in photographs}.
\newblock \bibinfo{journal}{\emph{arXiv preprint arXiv:1505.00110}}
  (\bibinfo{year}{2015}).
\newblock


\bibitem[\protect\citeauthoryear{{Cao}, {Hidalgo Martinez}, {Simon}, {Wei}, and
  {Sheikh}}{{Cao} et~al\mbox{.}}{2019}]%
        {8765346}
\bibfield{author}{\bibinfo{person}{Z. {Cao}}, \bibinfo{person}{G. {Hidalgo
  Martinez}}, \bibinfo{person}{T. {Simon}}, \bibinfo{person}{S. {Wei}}, {and}
  \bibinfo{person}{Y.~A. {Sheikh}}.} \bibinfo{year}{2019}\natexlab{}.
\newblock \showarticletitle{OpenPose: Realtime Multi-Person 2D Pose Estimation
  using Part Affinity Fields}.
\newblock \bibinfo{journal}{\emph{IEEE Transactions on Pattern Analysis and
  Machine Intelligence}} (\bibinfo{year}{2019}).
\newblock


\bibitem[\protect\citeauthoryear{Carneiro, Da~Silva, Del~Bue, and
  Costeira}{Carneiro et~al\mbox{.}}{2012}]%
        {carneiro2012artistic}
\bibfield{author}{\bibinfo{person}{Gustavo Carneiro},
  \bibinfo{person}{Nuno~Pinho Da~Silva}, \bibinfo{person}{Alessio Del~Bue},
  {and} \bibinfo{person}{Jo{\~a}o~Paulo Costeira}.}
  \bibinfo{year}{2012}\natexlab{}.
\newblock \showarticletitle{Artistic image classification: An analysis on the
  printart database}. In \bibinfo{booktitle}{\emph{European conference on
  computer vision}}. Springer, \bibinfo{pages}{143--157}.
\newblock


\bibitem[\protect\citeauthoryear{Cetinic, Lipic, and Grgic}{Cetinic
  et~al\mbox{.}}{2018}]%
        {cetinic2018fine}
\bibfield{author}{\bibinfo{person}{Eva Cetinic}, \bibinfo{person}{Tomislav
  Lipic}, {and} \bibinfo{person}{Sonja Grgic}.}
  \bibinfo{year}{2018}\natexlab{}.
\newblock \showarticletitle{Fine-tuning convolutional neural networks for fine
  art classification}.
\newblock \bibinfo{journal}{\emph{Expert Systems with Applications}}
  \bibinfo{volume}{114} (\bibinfo{year}{2018}), \bibinfo{pages}{107--118}.
\newblock


\bibitem[\protect\citeauthoryear{Cetinic, Lipic, and Grgic}{Cetinic
  et~al\mbox{.}}{2019}]%
        {cetinic2019deep}
\bibfield{author}{\bibinfo{person}{Eva Cetinic}, \bibinfo{person}{Tomislav
  Lipic}, {and} \bibinfo{person}{Sonja Grgic}.}
  \bibinfo{year}{2019}\natexlab{}.
\newblock \showarticletitle{A Deep Learning Perspective on Beauty, Sentiment,
  and Remembrance of Art}.
\newblock \bibinfo{journal}{\emph{IEEE Access}}  \bibinfo{volume}{7}
  (\bibinfo{year}{2019}), \bibinfo{pages}{73694--73710}.
\newblock


\bibitem[\protect\citeauthoryear{Chattopadhay, Sarkar, Howlader, and
  Balasubramanian}{Chattopadhay et~al\mbox{.}}{2018}]%
        {chattopadhay2018grad}
\bibfield{author}{\bibinfo{person}{Aditya Chattopadhay},
  \bibinfo{person}{Anirban Sarkar}, \bibinfo{person}{Prantik Howlader}, {and}
  \bibinfo{person}{Vineeth~N Balasubramanian}.}
  \bibinfo{year}{2018}\natexlab{}.
\newblock \showarticletitle{Grad-cam++: Generalized gradient-based visual
  explanations for deep convolutional networks}. In
  \bibinfo{booktitle}{\emph{2018 IEEE Winter Conference on Applications of
  Computer Vision (WACV)}}. IEEE, \bibinfo{pages}{839--847}.
\newblock


\bibitem[\protect\citeauthoryear{Chen, Chen, Zou, Huang, and Li}{Chen
  et~al\mbox{.}}{2017}]%
        {chen2017multi}
\bibfield{author}{\bibinfo{person}{Long Chen}, \bibinfo{person}{Jianda Chen},
  \bibinfo{person}{Qin Zou}, \bibinfo{person}{Kai Huang}, {and}
  \bibinfo{person}{Qingquan Li}.} \bibinfo{year}{2017}\natexlab{}.
\newblock \showarticletitle{Multi-view feature combination for ancient
  paintings chronological classification}.
\newblock \bibinfo{journal}{\emph{Journal on Computing and Cultural Heritage
  (JOCCH)}} \bibinfo{volume}{10}, \bibinfo{number}{2} (\bibinfo{year}{2017}),
  \bibinfo{pages}{1--15}.
\newblock


\bibitem[\protect\citeauthoryear{Chu and Wu}{Chu and Wu}{2016}]%
        {chu2016deep}
\bibfield{author}{\bibinfo{person}{Wei-Ta Chu} {and} \bibinfo{person}{Yi-Ling
  Wu}.} \bibinfo{year}{2016}\natexlab{}.
\newblock \showarticletitle{Deep correlation features for image style
  classification}. In \bibinfo{booktitle}{\emph{Proceedings of the 24th ACM
  international conference on Multimedia}}. \bibinfo{pages}{402--406}.
\newblock


\bibitem[\protect\citeauthoryear{Chu and Wu}{Chu and Wu}{2018}]%
        {chu2018image}
\bibfield{author}{\bibinfo{person}{Wei-Ta Chu} {and} \bibinfo{person}{Yi-Ling
  Wu}.} \bibinfo{year}{2018}\natexlab{}.
\newblock \showarticletitle{Image style classification based on learnt deep
  correlation features}.
\newblock \bibinfo{journal}{\emph{IEEE Transactions on Multimedia}}
  \bibinfo{volume}{20}, \bibinfo{number}{9} (\bibinfo{year}{2018}),
  \bibinfo{pages}{2491--2502}.
\newblock


\bibitem[\protect\citeauthoryear{Couprie}{Couprie}{1983}]%
        {couprie1983iconclass}
\bibfield{author}{\bibinfo{person}{Leendert~D Couprie}.}
  \bibinfo{year}{1983}\natexlab{}.
\newblock \showarticletitle{Iconclass: an iconographic classification system}.
\newblock \bibinfo{journal}{\emph{Art Libraries Journal}} \bibinfo{volume}{8},
  \bibinfo{number}{2} (\bibinfo{year}{1983}), \bibinfo{pages}{32--49}.
\newblock


\bibitem[\protect\citeauthoryear{Crowley and Zisserman}{Crowley and
  Zisserman}{2014a}]%
        {crowley2014state}
\bibfield{author}{\bibinfo{person}{Elliot Crowley} {and}
  \bibinfo{person}{Andrew Zisserman}.} \bibinfo{year}{2014}\natexlab{a}.
\newblock \showarticletitle{The State of the Art: Object Retrieval in Paintings
  using Discriminative Regions.}. In \bibinfo{booktitle}{\emph{BMVC}}.
\newblock


\bibitem[\protect\citeauthoryear{Crowley and Zisserman}{Crowley and
  Zisserman}{2014b}]%
        {BMVC.28.38}
\bibfield{author}{\bibinfo{person}{Elliot Crowley} {and}
  \bibinfo{person}{Andrew Zisserman}.} \bibinfo{year}{2014}\natexlab{b}.
\newblock \showarticletitle{The State of the Art: Object Retrieval in Paintings
  using Discriminative Regions}. In \bibinfo{booktitle}{\emph{Proceedings of
  the British Machine Vision Conference}}. \bibinfo{publisher}{BMVA Press}.
\newblock


\bibitem[\protect\citeauthoryear{Crowley and Zisserman}{Crowley and
  Zisserman}{2013}]%
        {crowley2013gods}
\bibfield{author}{\bibinfo{person}{Elliot~J Crowley} {and}
  \bibinfo{person}{Andrew Zisserman}.} \bibinfo{year}{2013}\natexlab{}.
\newblock \showarticletitle{Of gods and goats: Weakly supervised learning of
  figurative art}.
\newblock \bibinfo{journal}{\emph{learning}}  \bibinfo{volume}{8}
  (\bibinfo{year}{2013}), \bibinfo{pages}{14}.
\newblock


\bibitem[\protect\citeauthoryear{Crowley and Zisserman}{Crowley and
  Zisserman}{2016}]%
        {crowley2016art}
\bibfield{author}{\bibinfo{person}{Elliot~J Crowley} {and}
  \bibinfo{person}{Andrew Zisserman}.} \bibinfo{year}{2016}\natexlab{}.
\newblock \showarticletitle{The art of detection}. In
  \bibinfo{booktitle}{\emph{European Conference on Computer Vision}}. Springer,
  \bibinfo{pages}{721--737}.
\newblock


\bibitem[\protect\citeauthoryear{Deng, Dong, Socher, Li, Li, and Fei-Fei}{Deng
  et~al\mbox{.}}{2009}]%
        {deng2009imagenet}
\bibfield{author}{\bibinfo{person}{Jia Deng}, \bibinfo{person}{Wei Dong},
  \bibinfo{person}{Richard Socher}, \bibinfo{person}{Li-Jia Li},
  \bibinfo{person}{Kai Li}, {and} \bibinfo{person}{Li Fei-Fei}.}
  \bibinfo{year}{2009}\natexlab{}.
\newblock \showarticletitle{Imagenet: A large-scale hierarchical image
  database}. In \bibinfo{booktitle}{\emph{2009 IEEE conference on computer
  vision and pattern recognition}}. Ieee, \bibinfo{pages}{248--255}.
\newblock


\bibitem[\protect\citeauthoryear{Elgammal, Kang, and Den~Leeuw}{Elgammal
  et~al\mbox{.}}{2018}]%
        {elgammal2018picasso}
\bibfield{author}{\bibinfo{person}{Ahmed Elgammal}, \bibinfo{person}{Yan Kang},
  {and} \bibinfo{person}{Milko Den~Leeuw}.} \bibinfo{year}{2018}\natexlab{}.
\newblock \showarticletitle{Picasso, matisse, or a fake? Automated analysis of
  drawings at the stroke level for attribution and authentication}. In
  \bibinfo{booktitle}{\emph{Thirty-Second AAAI Conference on Artificial
  Intelligence}}.
\newblock


\bibitem[\protect\citeauthoryear{Everingham, Eslami, Van~Gool, Williams, Winn,
  and Zisserman}{Everingham et~al\mbox{.}}{2015}]%
        {everingham2015pascal}
\bibfield{author}{\bibinfo{person}{Mark Everingham}, \bibinfo{person}{SM~Ali
  Eslami}, \bibinfo{person}{Luc Van~Gool}, \bibinfo{person}{Christopher~KI
  Williams}, \bibinfo{person}{John Winn}, {and} \bibinfo{person}{Andrew
  Zisserman}.} \bibinfo{year}{2015}\natexlab{}.
\newblock \showarticletitle{The pascal visual object classes challenge: A
  retrospective}.
\newblock \bibinfo{journal}{\emph{International journal of computer vision}}
  \bibinfo{volume}{111}, \bibinfo{number}{1} (\bibinfo{year}{2015}),
  \bibinfo{pages}{98--136}.
\newblock


\bibitem[\protect\citeauthoryear{Falomir, Museros, Sanz, and
  Gonzalez-Abril}{Falomir et~al\mbox{.}}{2018}]%
        {falomir2018categorizing}
\bibfield{author}{\bibinfo{person}{Zoe Falomir}, \bibinfo{person}{Lled{\'o}
  Museros}, \bibinfo{person}{Ismael Sanz}, {and} \bibinfo{person}{Luis
  Gonzalez-Abril}.} \bibinfo{year}{2018}\natexlab{}.
\newblock \showarticletitle{Categorizing paintings in art styles based on
  qualitative color descriptors, quantitative global features and machine
  learning (QArt-Learn)}.
\newblock \bibinfo{journal}{\emph{Expert Systems with Applications}}
  \bibinfo{volume}{97} (\bibinfo{year}{2018}), \bibinfo{pages}{83--94}.
\newblock


\bibitem[\protect\citeauthoryear{Florea, Badea, Florea, and Vertan}{Florea
  et~al\mbox{.}}{2017}]%
        {florea2017domain}
\bibfield{author}{\bibinfo{person}{Corneliu Florea}, \bibinfo{person}{Mihai
  Badea}, \bibinfo{person}{Laura Florea}, {and} \bibinfo{person}{Constantin
  Vertan}.} \bibinfo{year}{2017}\natexlab{}.
\newblock \showarticletitle{Domain transfer for delving into deep networks
  capacity to de-abstract art}. In \bibinfo{booktitle}{\emph{Scandinavian
  Conference on Image Analysis}}. Springer, \bibinfo{pages}{337--349}.
\newblock


\bibitem[\protect\citeauthoryear{Florea, Condorovici, Vertan, Butnaru, Florea,
  and Vr{\^a}nceanu}{Florea et~al\mbox{.}}{2016}]%
        {florea2016pandora}
\bibfield{author}{\bibinfo{person}{Corneliu Florea},
  \bibinfo{person}{R{\u{a}}zvan Condorovici}, \bibinfo{person}{Constantin
  Vertan}, \bibinfo{person}{Raluca Butnaru}, \bibinfo{person}{Laura Florea},
  {and} \bibinfo{person}{Ruxandra Vr{\^a}nceanu}.}
  \bibinfo{year}{2016}\natexlab{}.
\newblock \showarticletitle{Pandora: Description of a painting database for art
  movement recognition with baselines and perspectives}. In
  \bibinfo{booktitle}{\emph{2016 24th European Signal Processing Conference
  (EUSIPCO)}}. IEEE, \bibinfo{pages}{918--922}.
\newblock


\bibitem[\protect\citeauthoryear{Gao, Shan, and Li}{Gao et~al\mbox{.}}{2015}]%
        {gao2015adaptive}
\bibfield{author}{\bibinfo{person}{Zhi Gao}, \bibinfo{person}{Mo Shan}, {and}
  \bibinfo{person}{Qingquan Li}.} \bibinfo{year}{2015}\natexlab{}.
\newblock \showarticletitle{Adaptive sparse representation for analyzing
  artistic style of paintings}.
\newblock \bibinfo{journal}{\emph{Journal on Computing and Cultural Heritage
  (JOCCH)}} \bibinfo{volume}{8}, \bibinfo{number}{4} (\bibinfo{year}{2015}),
  \bibinfo{pages}{1--15}.
\newblock


\bibitem[\protect\citeauthoryear{Ginosar, Haas, Brown, and Malik}{Ginosar
  et~al\mbox{.}}{2014}]%
        {ginosar2014detecting}
\bibfield{author}{\bibinfo{person}{Shiry Ginosar}, \bibinfo{person}{Daniel
  Haas}, \bibinfo{person}{Timothy Brown}, {and} \bibinfo{person}{Jitendra
  Malik}.} \bibinfo{year}{2014}\natexlab{}.
\newblock \showarticletitle{Detecting people in cubist art}. In
  \bibinfo{booktitle}{\emph{European Conference on Computer Vision}}. Springer,
  \bibinfo{pages}{101--116}.
\newblock


\bibitem[\protect\citeauthoryear{Girshick}{Girshick}{2015}]%
        {girshick2015fast}
\bibfield{author}{\bibinfo{person}{Ross Girshick}.}
  \bibinfo{year}{2015}\natexlab{}.
\newblock \showarticletitle{Fast r-cnn}. In
  \bibinfo{booktitle}{\emph{Proceedings of the IEEE international conference on
  computer vision}}. \bibinfo{pages}{1440--1448}.
\newblock


\bibitem[\protect\citeauthoryear{Gonthier, Gousseau, Ladjal, and
  Bonfait}{Gonthier et~al\mbox{.}}{2018}]%
        {gonthier2018weakly}
\bibfield{author}{\bibinfo{person}{Nicolas Gonthier}, \bibinfo{person}{Yann
  Gousseau}, \bibinfo{person}{Said Ladjal}, {and} \bibinfo{person}{Olivier
  Bonfait}.} \bibinfo{year}{2018}\natexlab{}.
\newblock \showarticletitle{Weakly Supervised Object Detection in Artworks}. In
  \bibinfo{booktitle}{\emph{Proceedings of the European Conference on Computer
  Vision (ECCV)}}. \bibinfo{pages}{0--0}.
\newblock


\bibitem[\protect\citeauthoryear{Gonthier, Ladjal, and Gousseau}{Gonthier
  et~al\mbox{.}}{2020}]%
        {gonthier2020multiple}
\bibfield{author}{\bibinfo{person}{Nicolas Gonthier},
  \bibinfo{person}{Sa{\"\i}d Ladjal}, {and} \bibinfo{person}{Yann Gousseau}.}
  \bibinfo{year}{2020}\natexlab{}.
\newblock \showarticletitle{Multiple instance learning on deep features for
  weakly supervised object detection with extreme domain shifts}.
\newblock \bibinfo{journal}{\emph{arXiv preprint arXiv:2008.01178}}
  (\bibinfo{year}{2020}).
\newblock


\bibitem[\protect\citeauthoryear{He, Zhang, Ren, and Sun}{He
  et~al\mbox{.}}{2016}]%
        {he2016deep}
\bibfield{author}{\bibinfo{person}{Kaiming He}, \bibinfo{person}{Xiangyu
  Zhang}, \bibinfo{person}{Shaoqing Ren}, {and} \bibinfo{person}{Jian Sun}.}
  \bibinfo{year}{2016}\natexlab{}.
\newblock \showarticletitle{Deep residual learning for image recognition}. In
  \bibinfo{booktitle}{\emph{Proceedings of the IEEE conference on computer
  vision and pattern recognition}}. \bibinfo{pages}{770--778}.
\newblock


\bibitem[\protect\citeauthoryear{Heinz-Mohr}{Heinz-Mohr}{1971}]%
        {heinz71}
\bibfield{author}{\bibinfo{person}{Gerd Heinz-Mohr}.}
  \bibinfo{year}{1971}\natexlab{}.
\newblock \bibinfo{booktitle}{\emph{Dictionary of Symbols, Images and Signs of
  Christian Art}}.
\newblock 320 pages.
\newblock


\bibitem[\protect\citeauthoryear{Huang, Zhong, and Xiao}{Huang
  et~al\mbox{.}}{2017}]%
        {huang2017fine}
\bibfield{author}{\bibinfo{person}{Xingsheng Huang}, \bibinfo{person}{Sheng-hua
  Zhong}, {and} \bibinfo{person}{Zhijiao Xiao}.}
  \bibinfo{year}{2017}\natexlab{}.
\newblock \showarticletitle{Fine-art painting classification via two-channel
  deep residual network}. In \bibinfo{booktitle}{\emph{Pacific Rim Conference
  on Multimedia}}. Springer, \bibinfo{pages}{79--88}.
\newblock


\bibitem[\protect\citeauthoryear{Johnson, Hendriks, Berezhnoy, Brevdo, Hughes,
  Daubechies, Li, Postma, and Wang}{Johnson et~al\mbox{.}}{2008}]%
        {johnson2008image}
\bibfield{author}{\bibinfo{person}{C~Richard Johnson}, \bibinfo{person}{Ella
  Hendriks}, \bibinfo{person}{Igor~J Berezhnoy}, \bibinfo{person}{Eugene
  Brevdo}, \bibinfo{person}{Shannon~M Hughes}, \bibinfo{person}{Ingrid
  Daubechies}, \bibinfo{person}{Jia Li}, \bibinfo{person}{Eric Postma}, {and}
  \bibinfo{person}{James~Z Wang}.} \bibinfo{year}{2008}\natexlab{}.
\newblock \showarticletitle{Image processing for artist identification}.
\newblock \bibinfo{journal}{\emph{IEEE Signal Processing Magazine}}
  \bibinfo{volume}{25}, \bibinfo{number}{4} (\bibinfo{year}{2008}),
  \bibinfo{pages}{37--48}.
\newblock


\bibitem[\protect\citeauthoryear{Kang, Shim, and Yoon}{Kang
  et~al\mbox{.}}{2018}]%
        {kang2018method}
\bibfield{author}{\bibinfo{person}{Dongwann Kang}, \bibinfo{person}{Hyounoh
  Shim}, {and} \bibinfo{person}{Kyunghyun Yoon}.}
  \bibinfo{year}{2018}\natexlab{}.
\newblock \showarticletitle{A method for extracting emotion using colors
  comprise the painting image}.
\newblock \bibinfo{journal}{\emph{Multimedia Tools and Applications}}
  \bibinfo{volume}{77}, \bibinfo{number}{4} (\bibinfo{year}{2018}),
  \bibinfo{pages}{4985--5002}.
\newblock


\bibitem[\protect\citeauthoryear{Karayev, Trentacoste, Han, Agarwala, Darrell,
  Hertzmann, and Winnemoeller}{Karayev et~al\mbox{.}}{2013}]%
        {karayev2013recognizing}
\bibfield{author}{\bibinfo{person}{Sergey Karayev}, \bibinfo{person}{Matthew
  Trentacoste}, \bibinfo{person}{Helen Han}, \bibinfo{person}{Aseem Agarwala},
  \bibinfo{person}{Trevor Darrell}, \bibinfo{person}{Aaron Hertzmann}, {and}
  \bibinfo{person}{Holger Winnemoeller}.} \bibinfo{year}{2013}\natexlab{}.
\newblock \showarticletitle{Recognizing image style}.
\newblock \bibinfo{journal}{\emph{arXiv preprint arXiv:1311.3715}}
  (\bibinfo{year}{2013}).
\newblock


\bibitem[\protect\citeauthoryear{Keren}{Keren}{2002}]%
        {keren2002painter}
\bibfield{author}{\bibinfo{person}{Daniel Keren}.}
  \bibinfo{year}{2002}\natexlab{}.
\newblock \showarticletitle{Painter identification using local features and
  naive bayes}. In \bibinfo{booktitle}{\emph{Object recognition supported by
  user interaction for service robots}}, Vol.~\bibinfo{volume}{2}. IEEE,
  \bibinfo{pages}{474--477}.
\newblock


\bibitem[\protect\citeauthoryear{Khan, Beigpour, Weijer, and Felsberg}{Khan
  et~al\mbox{.}}{2014}]%
        {painting91}
\bibfield{author}{\bibinfo{person}{Fahad Khan}, \bibinfo{person}{Shida
  Beigpour}, \bibinfo{person}{Joost Weijer}, {and} \bibinfo{person}{Michael
  Felsberg}.} \bibinfo{year}{2014}\natexlab{}.
\newblock \showarticletitle{Painting-91: A large scale database for
  computational painting categorization}.
\newblock \bibinfo{journal}{\emph{Machine Vision and Applications}}
  \bibinfo{volume}{25} (\bibinfo{date}{08} \bibinfo{year}{2014}),
  \bibinfo{pages}{1385--1397}.
\newblock
\urldef\tempurl%
\url{https://doi.org/10.1007/s00138-014-0621-6}
\showDOI{\tempurl}


\bibitem[\protect\citeauthoryear{Krizhevsky, Sutskever, and Hinton}{Krizhevsky
  et~al\mbox{.}}{2012}]%
        {krizhevsky2012imagenet}
\bibfield{author}{\bibinfo{person}{Alex Krizhevsky}, \bibinfo{person}{Ilya
  Sutskever}, {and} \bibinfo{person}{Geoffrey~E Hinton}.}
  \bibinfo{year}{2012}\natexlab{}.
\newblock \showarticletitle{Imagenet classification with deep convolutional
  neural networks}. In \bibinfo{booktitle}{\emph{Advances in neural information
  processing systems}}. \bibinfo{pages}{1097--1105}.
\newblock


\bibitem[\protect\citeauthoryear{Lanzi and Lanzi}{Lanzi and Lanzi}{2004}]%
        {Lanzi04}
\bibfield{author}{\bibinfo{person}{Fernando Lanzi} {and} \bibinfo{person}{Gioia
  Lanzi}.} \bibinfo{year}{2004}\natexlab{}.
\newblock \bibinfo{booktitle}{\emph{Saints and their Symbols: Recognizing
  Saints in Art and in Popular Images}}.
\newblock 237 pages.
\newblock


\bibitem[\protect\citeauthoryear{Lecoutre, Negrevergne, and Yger}{Lecoutre
  et~al\mbox{.}}{2017}]%
        {lecoutre2017recognizing}
\bibfield{author}{\bibinfo{person}{Adrian Lecoutre}, \bibinfo{person}{Benjamin
  Negrevergne}, {and} \bibinfo{person}{Florian Yger}.}
  \bibinfo{year}{2017}\natexlab{}.
\newblock \showarticletitle{Recognizing art style automatically in painting
  with deep learning}. In \bibinfo{booktitle}{\emph{Asian conference on machine
  learning}}. \bibinfo{pages}{327--342}.
\newblock


\bibitem[\protect\citeauthoryear{Li and Chen}{Li and Chen}{2009}]%
        {li2009aesthetic}
\bibfield{author}{\bibinfo{person}{Congcong Li} {and} \bibinfo{person}{Tsuhan
  Chen}.} \bibinfo{year}{2009}\natexlab{}.
\newblock \showarticletitle{Aesthetic visual quality assessment of paintings}.
\newblock \bibinfo{journal}{\emph{IEEE Journal of selected topics in Signal
  Processing}} \bibinfo{volume}{3}, \bibinfo{number}{2} (\bibinfo{year}{2009}),
  \bibinfo{pages}{236--252}.
\newblock


\bibitem[\protect\citeauthoryear{Lin, Maire, Belongie, Hays, Perona, Ramanan,
  Doll{\'a}r, and Zitnick}{Lin et~al\mbox{.}}{2014}]%
        {lin2014microsoft}
\bibfield{author}{\bibinfo{person}{Tsung-Yi Lin}, \bibinfo{person}{Michael
  Maire}, \bibinfo{person}{Serge Belongie}, \bibinfo{person}{James Hays},
  \bibinfo{person}{Pietro Perona}, \bibinfo{person}{Deva Ramanan},
  \bibinfo{person}{Piotr Doll{\'a}r}, {and} \bibinfo{person}{C~Lawrence
  Zitnick}.} \bibinfo{year}{2014}\natexlab{}.
\newblock \showarticletitle{Microsoft coco: Common objects in context}. In
  \bibinfo{booktitle}{\emph{European conference on computer vision}}. Springer,
  \bibinfo{pages}{740--755}.
\newblock


\bibitem[\protect\citeauthoryear{Mao, Cheung, and She}{Mao
  et~al\mbox{.}}{2017}]%
        {mao2017deepart}
\bibfield{author}{\bibinfo{person}{Hui Mao}, \bibinfo{person}{Ming Cheung},
  {and} \bibinfo{person}{James She}.} \bibinfo{year}{2017}\natexlab{}.
\newblock \showarticletitle{Deepart: Learning joint representations of visual
  arts}. In \bibinfo{booktitle}{\emph{Proceedings of the 25th ACM international
  conference on Multimedia}}. ACM, \bibinfo{pages}{1183--1191}.
\newblock


\bibitem[\protect\citeauthoryear{Mao, She, and Cheung}{Mao
  et~al\mbox{.}}{2019}]%
        {mao2019visual}
\bibfield{author}{\bibinfo{person}{Hui Mao}, \bibinfo{person}{James She}, {and}
  \bibinfo{person}{Ming Cheung}.} \bibinfo{year}{2019}\natexlab{}.
\newblock \showarticletitle{Visual Arts Search on Mobile Devices}.
\newblock \bibinfo{journal}{\emph{ACM Transactions on Multimedia Computing,
  Communications, and Applications (TOMM)}} \bibinfo{volume}{15},
  \bibinfo{number}{2s} (\bibinfo{year}{2019}), \bibinfo{pages}{60}.
\newblock


\bibitem[\protect\citeauthoryear{Mensink and Van~Gemert}{Mensink and
  Van~Gemert}{2014}]%
        {mensink2014rijksmuseum}
\bibfield{author}{\bibinfo{person}{Thomas Mensink} {and} \bibinfo{person}{Jan
  Van~Gemert}.} \bibinfo{year}{2014}\natexlab{}.
\newblock \showarticletitle{The rijksmuseum challenge: Museum-centered visual
  recognition}. In \bibinfo{booktitle}{\emph{Proceedings of International
  Conference on Multimedia Retrieval}}. ACM, \bibinfo{pages}{451}.
\newblock


\bibitem[\protect\citeauthoryear{Mensink and van Gemert}{Mensink and van
  Gemert}{2014}]%
        {MensinkICMIR2014}
\bibfield{author}{\bibinfo{person}{T.~E.~J. Mensink} {and}
  \bibinfo{person}{J.~C. van Gemert}.} \bibinfo{year}{2014}\natexlab{}.
\newblock \showarticletitle{The Rijksmuseum Challenge: Museum-Centered Visual
  Recognition}. In \bibinfo{booktitle}{\emph{ACM International Conference on
  Multimedia Retrieval}}.
\newblock
\urldef\tempurl%
\url{https://ivi.fnwi.uva.nl/isis/publications/2014/MensinkICMIR2014}
\showURL{%
\tempurl}


\bibitem[\protect\citeauthoryear{Omeiza, Speakman, Cintas, and
  Weldermariam}{Omeiza et~al\mbox{.}}{2019}]%
        {omeiza2019smooth}
\bibfield{author}{\bibinfo{person}{Daniel Omeiza}, \bibinfo{person}{Skyler
  Speakman}, \bibinfo{person}{Celia Cintas}, {and} \bibinfo{person}{Komminist
  Weldermariam}.} \bibinfo{year}{2019}\natexlab{}.
\newblock \showarticletitle{Smooth grad-cam++: An enhanced inference level
  visualization technique for deep convolutional neural network models}.
\newblock \bibinfo{journal}{\emph{arXiv preprint arXiv:1908.01224}}
  (\bibinfo{year}{2019}).
\newblock


\bibitem[\protect\citeauthoryear{Pan and Yang}{Pan and Yang}{2009}]%
        {pan2009survey}
\bibfield{author}{\bibinfo{person}{Sinno~Jialin Pan} {and}
  \bibinfo{person}{Qiang Yang}.} \bibinfo{year}{2009}\natexlab{}.
\newblock \showarticletitle{A survey on transfer learning}.
\newblock \bibinfo{journal}{\emph{IEEE Transactions on knowledge and data
  engineering}} \bibinfo{volume}{22}, \bibinfo{number}{10}
  (\bibinfo{year}{2009}), \bibinfo{pages}{1345--1359}.
\newblock


\bibitem[\protect\citeauthoryear{Panofsky}{Panofsky}{1939}]%
        {panofsky39}
\bibfield{author}{\bibinfo{person}{Erwin Panofsky}.}
  \bibinfo{year}{1939}\natexlab{}.
\newblock \bibinfo{booktitle}{\emph{Studies in Iconology: Humanistic Themes in
  the Art of the Renaissance}}.
\newblock 262 pages.
\newblock


\bibitem[\protect\citeauthoryear{Peng and Chen}{Peng and Chen}{2015}]%
        {peng2015cross}
\bibfield{author}{\bibinfo{person}{Kuan-Chuan Peng} {and}
  \bibinfo{person}{Tsuhan Chen}.} \bibinfo{year}{2015}\natexlab{}.
\newblock \showarticletitle{Cross-layer features in convolutional neural
  networks for generic classification tasks}. In \bibinfo{booktitle}{\emph{2015
  IEEE International Conference on Image Processing (ICIP)}}. IEEE,
  \bibinfo{pages}{3057--3061}.
\newblock


\bibitem[\protect\citeauthoryear{Peng and Chen}{Peng and Chen}{2016}]%
        {peng2016toward}
\bibfield{author}{\bibinfo{person}{Kuan-Chuan Peng} {and}
  \bibinfo{person}{Tsuhan Chen}.} \bibinfo{year}{2016}\natexlab{}.
\newblock \showarticletitle{Toward correlating and solving abstract tasks using
  convolutional neural networks}. In \bibinfo{booktitle}{\emph{2016 IEEE Winter
  Conference on Applications of Computer Vision (WACV)}}. IEEE,
  \bibinfo{pages}{1--9}.
\newblock


\bibitem[\protect\citeauthoryear{Puthenputhussery, Liu, and
  Liu}{Puthenputhussery et~al\mbox{.}}{2016a}]%
        {puthenputhussery2016color}
\bibfield{author}{\bibinfo{person}{Ajit Puthenputhussery},
  \bibinfo{person}{Qingfeng Liu}, {and} \bibinfo{person}{Chengjun Liu}.}
  \bibinfo{year}{2016}\natexlab{a}.
\newblock \showarticletitle{Color multi-fusion fisher vector feature for fine
  art painting categorization and influence analysis}. In
  \bibinfo{booktitle}{\emph{2016 IEEE Winter Conference on Applications of
  Computer Vision (WACV)}}. IEEE, \bibinfo{pages}{1--9}.
\newblock


\bibitem[\protect\citeauthoryear{Puthenputhussery, Liu, and
  Liu}{Puthenputhussery et~al\mbox{.}}{2016b}]%
        {puthenputhussery2016sparse}
\bibfield{author}{\bibinfo{person}{Ajit Puthenputhussery},
  \bibinfo{person}{Qingfeng Liu}, {and} \bibinfo{person}{Chengjun Liu}.}
  \bibinfo{year}{2016}\natexlab{b}.
\newblock \showarticletitle{Sparse representation based complete kernel
  marginal fisher analysis framework for computational art painting
  categorization}. In \bibinfo{booktitle}{\emph{European Conference on Computer
  Vision}}. Springer, \bibinfo{pages}{612--627}.
\newblock


\bibitem[\protect\citeauthoryear{Sabatelli, Kestemont, Daelemans, and
  Geurts}{Sabatelli et~al\mbox{.}}{2018}]%
        {sabatelli2018deep}
\bibfield{author}{\bibinfo{person}{Matthia Sabatelli}, \bibinfo{person}{Mike
  Kestemont}, \bibinfo{person}{Walter Daelemans}, {and} \bibinfo{person}{Pierre
  Geurts}.} \bibinfo{year}{2018}\natexlab{}.
\newblock \showarticletitle{Deep transfer learning for art classification
  problems}. In \bibinfo{booktitle}{\emph{Proceedings of the European
  Conference on Computer Vision (ECCV)}}. \bibinfo{pages}{0--0}.
\newblock


\bibitem[\protect\citeauthoryear{Saleh and Elgammal}{Saleh and
  Elgammal}{2015}]%
        {saleh2015large}
\bibfield{author}{\bibinfo{person}{Babak Saleh} {and} \bibinfo{person}{Ahmed
  Elgammal}.} \bibinfo{year}{2015}\natexlab{}.
\newblock \showarticletitle{Large-scale classification of fine-art paintings:
  Learning the right metric on the right feature}.
\newblock \bibinfo{journal}{\emph{arXiv preprint arXiv:1505.00855}}
  (\bibinfo{year}{2015}).
\newblock


\bibitem[\protect\citeauthoryear{Salimans, Goodfellow, Zaremba, Cheung,
  Radford, Chen, and Chen}{Salimans et~al\mbox{.}}{2016}]%
        {Salimans2016}
\bibfield{author}{\bibinfo{person}{Tim Salimans}, \bibinfo{person}{Ian
  Goodfellow}, \bibinfo{person}{Wojciech Zaremba}, \bibinfo{person}{Vicki
  Cheung}, \bibinfo{person}{Alec Radford}, \bibinfo{person}{Xi Chen}, {and}
  \bibinfo{person}{Xi Chen}.} \bibinfo{year}{2016}\natexlab{}.
\newblock \showarticletitle{Improved Techniques for Training GANs}.
\newblock In \bibinfo{booktitle}{\emph{Advances in Neural Information
  Processing Systems 29}}, \bibfield{editor}{\bibinfo{person}{D.~D. Lee},
  \bibinfo{person}{M.~Sugiyama}, \bibinfo{person}{U.~V. Luxburg},
  \bibinfo{person}{I.~Guyon}, {and} \bibinfo{person}{R.~Garnett}} (Eds.).
  \bibinfo{publisher}{Curran Associates, Inc.}, \bibinfo{pages}{2234--2242}.
\newblock
\urldef\tempurl%
\url{http://papers.nips.cc/paper/6125-improved-techniques-for-training-gans.pdf}
\showURL{%
\tempurl}


\bibitem[\protect\citeauthoryear{Selvaraju, Cogswell, Das, Vedantam, Parikh,
  and Batra}{Selvaraju et~al\mbox{.}}{2017}]%
        {selvaraju2017grad}
\bibfield{author}{\bibinfo{person}{Ramprasaath~R Selvaraju},
  \bibinfo{person}{Michael Cogswell}, \bibinfo{person}{Abhishek Das},
  \bibinfo{person}{Ramakrishna Vedantam}, \bibinfo{person}{Devi Parikh}, {and}
  \bibinfo{person}{Dhruv Batra}.} \bibinfo{year}{2017}\natexlab{}.
\newblock \showarticletitle{Grad-cam: Visual explanations from deep networks
  via gradient-based localization}. In \bibinfo{booktitle}{\emph{Proceedings of
  the IEEE international conference on computer vision}}.
  \bibinfo{pages}{618--626}.
\newblock


\bibitem[\protect\citeauthoryear{Shamir and Tarakhovsky}{Shamir and
  Tarakhovsky}{2012}]%
        {shamir2012computer}
\bibfield{author}{\bibinfo{person}{Lior Shamir} {and} \bibinfo{person}{Jane~A
  Tarakhovsky}.} \bibinfo{year}{2012}\natexlab{}.
\newblock \showarticletitle{Computer analysis of art}.
\newblock \bibinfo{journal}{\emph{Journal on Computing and Cultural Heritage
  (JOCCH)}} \bibinfo{volume}{5}, \bibinfo{number}{2} (\bibinfo{year}{2012}),
  \bibinfo{pages}{1--11}.
\newblock


\bibitem[\protect\citeauthoryear{Shen, Efros, and Mathieu}{Shen
  et~al\mbox{.}}{2019}]%
        {shen2019discovering}
\bibfield{author}{\bibinfo{person}{Xi Shen}, \bibinfo{person}{Alexei~A Efros},
  {and} \bibinfo{person}{Aubry Mathieu}.} \bibinfo{year}{2019}\natexlab{}.
\newblock \showarticletitle{Discovering Visual Patterns in Art Collections with
  Spatially-consistent Feature Learning}.
\newblock \bibinfo{journal}{\emph{arXiv preprint arXiv:1903.02678}}
  (\bibinfo{year}{2019}).
\newblock


\bibitem[\protect\citeauthoryear{Springenberg, Dosovitskiy, Brox, and
  Riedmiller}{Springenberg et~al\mbox{.}}{2014}]%
        {springenberg2014striving}
\bibfield{author}{\bibinfo{person}{Jost~Tobias Springenberg},
  \bibinfo{person}{Alexey Dosovitskiy}, \bibinfo{person}{Thomas Brox}, {and}
  \bibinfo{person}{Martin Riedmiller}.} \bibinfo{year}{2014}\natexlab{}.
\newblock \showarticletitle{Striving for simplicity: The all convolutional
  net}.
\newblock \bibinfo{journal}{\emph{arXiv preprint arXiv:1412.6806}}
  (\bibinfo{year}{2014}).
\newblock


\bibitem[\protect\citeauthoryear{Stefanini, Cornia, Baraldi, Corsini, and
  Cucchiara}{Stefanini et~al\mbox{.}}{2019}]%
        {stefanini2019artpedia}
\bibfield{author}{\bibinfo{person}{Matteo Stefanini}, \bibinfo{person}{Marcella
  Cornia}, \bibinfo{person}{Lorenzo Baraldi}, \bibinfo{person}{Massimiliano
  Corsini}, {and} \bibinfo{person}{Rita Cucchiara}.}
  \bibinfo{year}{2019}\natexlab{}.
\newblock \showarticletitle{Artpedia: A New Visual-Semantic Dataset with Visual
  and Contextual Sentences in the Artistic Domain}. In
  \bibinfo{booktitle}{\emph{20th International Conference on Image Analysis and
  Processing}}.
\newblock


\bibitem[\protect\citeauthoryear{Strezoski, Knoester, van Noord, and
  Worring}{Strezoski et~al\mbox{.}}{2020}]%
        {strezoski2020omnieyes}
\bibfield{author}{\bibinfo{person}{Gjorgji Strezoski}, \bibinfo{person}{Rogier
  Knoester}, \bibinfo{person}{Nanne van Noord}, {and} \bibinfo{person}{Marcel
  Worring}.} \bibinfo{year}{2020}\natexlab{}.
\newblock \showarticletitle{OmniEyes: Analysis and Synthesis of Artistically
  Painted Eyes}. In \bibinfo{booktitle}{\emph{International Conference on
  Multimedia Modeling}}. Springer, \bibinfo{pages}{628--641}.
\newblock


\bibitem[\protect\citeauthoryear{Strezoski and Worring}{Strezoski and
  Worring}{2017}]%
        {strezoski2017omniart}
\bibfield{author}{\bibinfo{person}{Gjorgji Strezoski} {and}
  \bibinfo{person}{Marcel Worring}.} \bibinfo{year}{2017}\natexlab{}.
\newblock \showarticletitle{Omniart: multi-task deep learning for artistic data
  analysis}.
\newblock \bibinfo{journal}{\emph{arXiv preprint arXiv:1708.00684}}
  (\bibinfo{year}{2017}).
\newblock


\bibitem[\protect\citeauthoryear{Strezoski and Worring}{Strezoski and
  Worring}{2018}]%
        {strezoski2018omniart}
\bibfield{author}{\bibinfo{person}{Gjorgji Strezoski} {and}
  \bibinfo{person}{Marcel Worring}.} \bibinfo{year}{2018}\natexlab{}.
\newblock \showarticletitle{Omniart: A large-scale artistic benchmark}.
\newblock \bibinfo{journal}{\emph{ACM Transactions on Multimedia Computing,
  Communications, and Applications (TOMM)}} \bibinfo{volume}{14},
  \bibinfo{number}{4} (\bibinfo{year}{2018}), \bibinfo{pages}{88}.
\newblock


\bibitem[\protect\citeauthoryear{Tan, Chan, Aguirre, and Tanaka}{Tan
  et~al\mbox{.}}{2016}]%
        {tan2016ceci}
\bibfield{author}{\bibinfo{person}{Wei~Ren Tan}, \bibinfo{person}{Chee~Seng
  Chan}, \bibinfo{person}{Hern{\'a}n~E Aguirre}, {and} \bibinfo{person}{Kiyoshi
  Tanaka}.} \bibinfo{year}{2016}\natexlab{}.
\newblock \showarticletitle{Ceci n'est pas une pipe: A deep convolutional
  network for fine-art paintings classification}. In
  \bibinfo{booktitle}{\emph{2016 IEEE international conference on image
  processing (ICIP)}}. IEEE, \bibinfo{pages}{3703--3707}.
\newblock


\bibitem[\protect\citeauthoryear{Westlake, Cai, and Hall}{Westlake
  et~al\mbox{.}}{2016}]%
        {westlake2016detecting}
\bibfield{author}{\bibinfo{person}{Nicholas Westlake},
  \bibinfo{person}{Hongping Cai}, {and} \bibinfo{person}{Peter Hall}.}
  \bibinfo{year}{2016}\natexlab{}.
\newblock \showarticletitle{Detecting people in artwork with CNNs}. In
  \bibinfo{booktitle}{\emph{European Conference on Computer Vision}}. Springer,
  \bibinfo{pages}{825--841}.
\newblock


\bibitem[\protect\citeauthoryear{Wilber, Fang, Jin, Hertzmann, Collomosse, and
  Belongie}{Wilber et~al\mbox{.}}{2017}]%
        {wilber2017bam}
\bibfield{author}{\bibinfo{person}{Michael~J Wilber}, \bibinfo{person}{Chen
  Fang}, \bibinfo{person}{Hailin Jin}, \bibinfo{person}{Aaron Hertzmann},
  \bibinfo{person}{John Collomosse}, {and} \bibinfo{person}{Serge Belongie}.}
  \bibinfo{year}{2017}\natexlab{}.
\newblock \showarticletitle{Bam! the behance artistic media dataset for
  recognition beyond photography}. In \bibinfo{booktitle}{\emph{Proceedings of
  the IEEE International Conference on Computer Vision}}.
  \bibinfo{pages}{1202--1211}.
\newblock


\bibitem[\protect\citeauthoryear{Yosinski, Clune, Bengio, and Lipson}{Yosinski
  et~al\mbox{.}}{2014}]%
        {yosinski2014transferable}
\bibfield{author}{\bibinfo{person}{Jason Yosinski}, \bibinfo{person}{Jeff
  Clune}, \bibinfo{person}{Yoshua Bengio}, {and} \bibinfo{person}{Hod Lipson}.}
  \bibinfo{year}{2014}\natexlab{}.
\newblock \showarticletitle{How transferable are features in deep neural
  networks?}. In \bibinfo{booktitle}{\emph{Advances in neural information
  processing systems}}. \bibinfo{pages}{3320--3328}.
\newblock


\bibitem[\protect\citeauthoryear{Zhong, Huang, and Xiao}{Zhong
  et~al\mbox{.}}{2020}]%
        {zhong2020fine}
\bibfield{author}{\bibinfo{person}{Sheng-hua Zhong}, \bibinfo{person}{Xingsheng
  Huang}, {and} \bibinfo{person}{Zhijiao Xiao}.}
  \bibinfo{year}{2020}\natexlab{}.
\newblock \showarticletitle{Fine-art painting classification via two-channel
  dual path networks}.
\newblock \bibinfo{journal}{\emph{International Journal of Machine Learning and
  Cybernetics}} \bibinfo{volume}{11}, \bibinfo{number}{1}
  (\bibinfo{year}{2020}), \bibinfo{pages}{137--152}.
\newblock


\bibitem[\protect\citeauthoryear{Zhou, Khosla, Lapedriza, Oliva, and
  Torralba}{Zhou et~al\mbox{.}}{2016}]%
        {zhou2016learning}
\bibfield{author}{\bibinfo{person}{Bolei Zhou}, \bibinfo{person}{Aditya
  Khosla}, \bibinfo{person}{Agata Lapedriza}, \bibinfo{person}{Aude Oliva},
  {and} \bibinfo{person}{Antonio Torralba}.} \bibinfo{year}{2016}\natexlab{}.
\newblock \showarticletitle{Learning deep features for discriminative
  localization}. In \bibinfo{booktitle}{\emph{Proceedings of the IEEE
  conference on computer vision and pattern recognition}}.
  \bibinfo{pages}{2921--2929}.
\newblock


\bibitem[\protect\citeauthoryear{Zujovic, Gandy, Friedman, Pardo, and
  Pappas}{Zujovic et~al\mbox{.}}{2009}]%
        {zujovic2009classifying}
\bibfield{author}{\bibinfo{person}{Jana Zujovic}, \bibinfo{person}{Lisa Gandy},
  \bibinfo{person}{Scott Friedman}, \bibinfo{person}{Bryan Pardo}, {and}
  \bibinfo{person}{Thrasyvoulos~N Pappas}.} \bibinfo{year}{2009}\natexlab{}.
\newblock \showarticletitle{Classifying paintings by artistic genre: An
  analysis of features \& classifiers}. In \bibinfo{booktitle}{\emph{2009 IEEE
  International Workshop on Multimedia Signal Processing}}. IEEE,
  \bibinfo{pages}{1--5}.
\newblock


\end{thebibliography}

\end{document}